\documentclass[conference]{IEEEtran}
\usepackage{times}
\usepackage{graphicx}

\usepackage{YJ_mystyle}

\usepackage{algorithm}
\usepackage[noend]{algpseudocode}

\theoremstyle{plain}\newtheorem{thm}{Theorem}
\theoremstyle{definition}\newtheorem{defn}{Definition}
\theoremstyle{plain}
\theoremstyle{plain}

\usepackage{todonotes}

\usepackage[numbers]{natbib}
\usepackage{multicol}
\usepackage[bookmarks=true]{hyperref}

\newcommand{\powerset}[1]{\mathcal{P}(#1)}
\newcommand{\BigO}[1]{\mathcal{O}(#1)}

\newcommand{\proc}[1]{\textsc{#1}}
\newcommand{\kw}[1]{\textbf{#1}}

\newcommand{\transpose}[1]{{#1}^\intercal}

\begin{document}

%\title{Sequence and motion planning for robotic spatial extrusion}
\title{Scalable and Probabilistically Complete Planning for Robotic Spatial Extrusion}

% You will get a Paper-ID when submitting a pdf file to the conference system
% \author{Paper 27}

% \author{\authorblockN{Caelan Reed Garrett} % \thanks{equal contribution}}, authorrefmark
% \authorblockA{
% Massachusetts Institute of Technology\\
% % MIT CSAIL\\
% % Cambridge, MA 02139\\
% %Email: 
% \href{mailto:caelan@csail.mit.edu}{caelan@csail.mit.edu}}
% \and
% \authorblockN{Yijiang Huang} %\footnotemark[1]}
% \authorblockA{
% Department of Architecture\\
% %Cambridge, MA 02139\\
% \href{mailto:yijiangh@mit.edu}{yijiangh@mit.edu}}
% \and
% \authorblockN{Tom\'as Lozano-P\'erez}
% \authorblockA{
% Massachusetts Institute of Technology\\
% %Cambridge, MA 02139\\
% \href{mailto:tlp@csail.mit.edu}{tlp@csail.mit.edu}}
% \and
% \authorblockN{Caitlin Tobin Mueller}
% \authorblockA{
% Department of Architecture\\
% %MIT Architecture\\
% %Cambridge, MA 02139\\
% \href{caitlinm@mit.edu}{caitlinm@mit.edu}}}

% Computer  Science and Artificial Intelligence Laboratory
% Caelan Reed Garrett, Yijiang Huang, Tomás Lozano-Pérez, Caitlin Tobin Mueller

% avoiding spaces at the end of the author lines is not a problem with
% conference papers because we don't use \thanks or \IEEEmembership

\author{\authorblockN{Caelan Reed Garrett\authorrefmark{2}\authorrefmark{1},
Yijiang Huang\authorrefmark{3}\authorrefmark{1},
Tom\'as Lozano-P\'erez\authorrefmark{2} and
Caitlin Tobin Mueller\authorrefmark{3}}
\authorblockA{\authorrefmark{2}Computer Science and Artificial Intelligence Laboratory, %//
Massachusetts Institute of Technology \\
%Cambridge, Massachusetts 02139\\ 
Email: \{caelan,tlp\}@csail.mit.edu}
\authorblockA{\authorrefmark{3}Department of Architecture, %\\
Massachusetts Institute of Technology \\
Email: \{yijiangh,caitlinm\}@mit.edu}
\authorblockA{\begin{small}\authorrefmark{1}Authors contributed equally\end{small}}}

\maketitle

\begin{abstract}
    There is increasing demand for automated systems that can fabricate 3D structures.
    % in construction 
    Robotic spatial extrusion has become an attractive alternative to traditional layer-based 3D printing due to a manipulator's flexibility to print large, directionally-dependent structures.
    % anisotropic
    %scale, flexiblity, DOFs, different structures, quickly
    However, existing extrusion planning algorithms require a substantial amount of human input, do not scale to large instances, and lack theoretical guarantees.
    In this work, we present a rigorous formalization of robotic spatial extrusion planning and provide several efficient and probabilistically complete planning algorithms.
    The key planning challenge is, throughout the printing process, satisfying both {\em stiffness} constraints that limit the deformation of the structure and {\em geometric} constraints that ensure the robot does not collide with the structure.
    We show that, although these constraints often conflict with each other, a greedy backward state-space search guided by a stiffness-aware heuristic is able to successfully balance both constraints.
    %dichotomy
    We empirically compare our methods on a benchmark of over 40 simulated extrusion problems.
    Finally, we apply our approach to 3 real-world extrusion problems.
\end{abstract}

\IEEEpeerreviewmaketitle

% https://docs.google.com/document/d/1yHloWNPAF_gmaJormVTEeWp090z6Vqe8MO70H_gRPCs/edit
% https://www.mendeley.com/community/construction-9/
% http://rss2019.informatik.uni-freiburg.de/information/authorinfo/
% https://roboticsconference.org/
% https://www.researchgate.net/profile/Yijiang_Huang2/publication/328040374_Automated_sequence_and_motion_planning_for_robotic_spatial_extrusion_of_3D_trusses/links/5ca89b76a6fdcca26d01a3d1/Automated-sequence-and-motion-planning-for-robotic-spatial-extrusion-of-3D-trusses.pdf
% https://github.com/yijiangh/conmech/blob/master/tests/notebook_demo/demo.ipynb

\section{Introduction}

%\YJ{talk about why frame structures are useful and practical. Explain the sources of the printed artifacts shown in this paper}
% The advancement of additive manufacturing offers unprecedented means for achieving geometrical complexity across different scales - enabling mass customization and endless flexibility in fabricating discrete and continuum shapes. 

\begin{figure}[!ht]
  \centering
  \includegraphics[width=0.445\columnwidth]{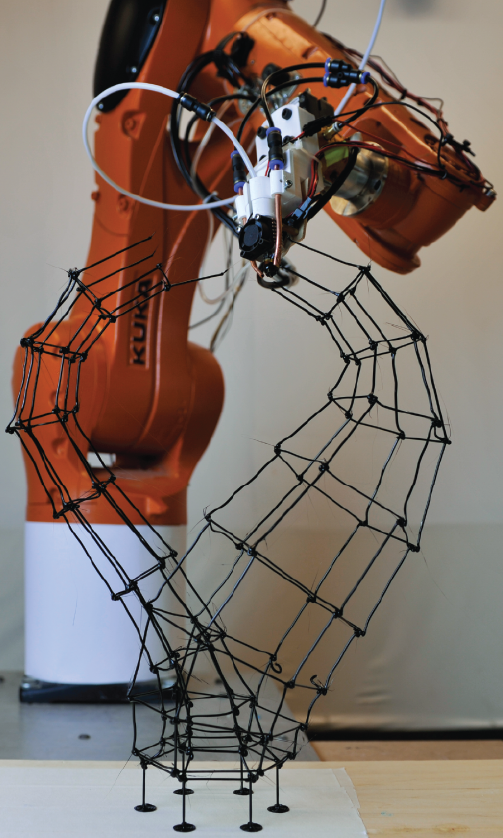}
  \includegraphics[width=0.535\columnwidth]{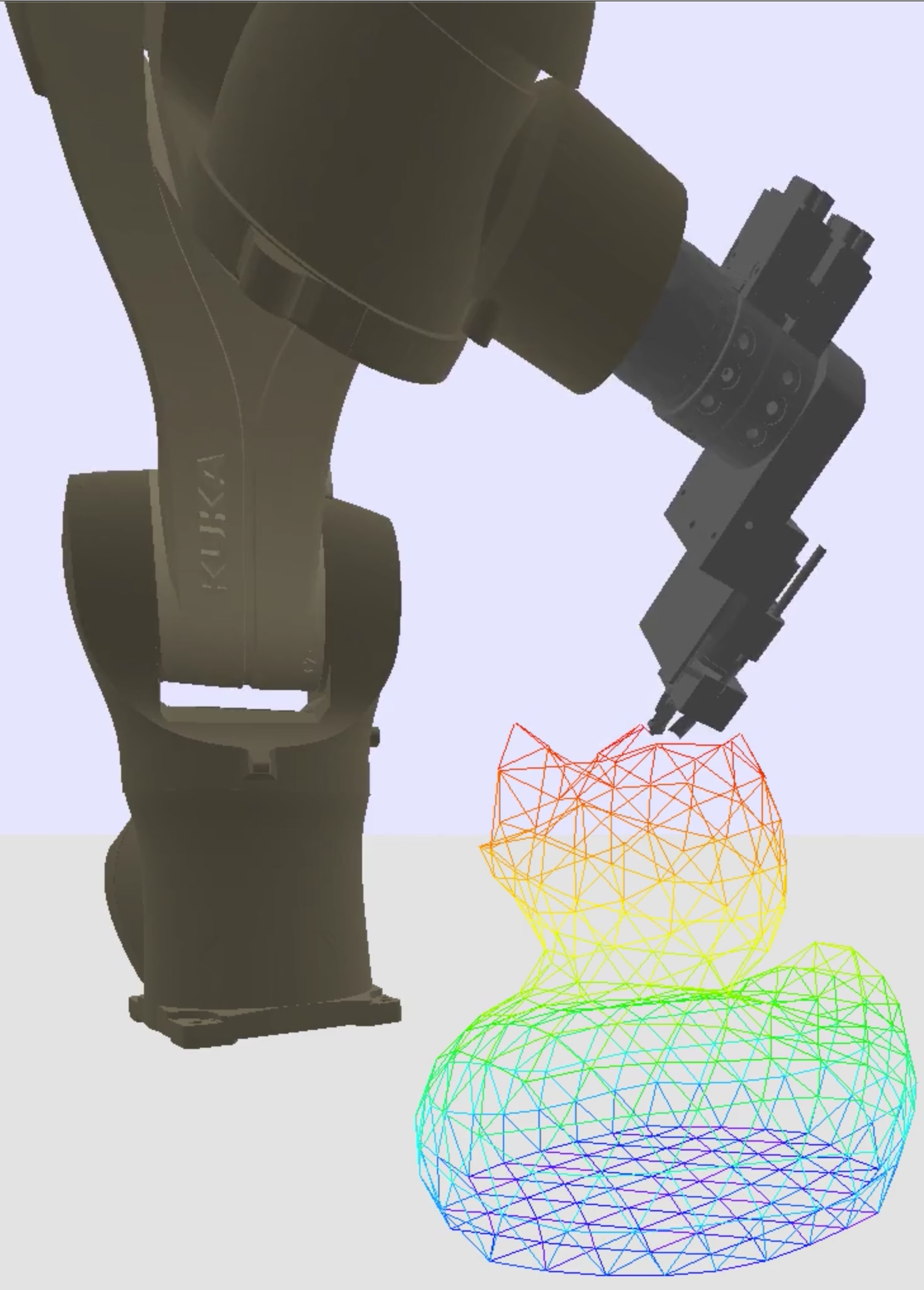}
  \caption{{\em Left}: Klein bottle (246 elements). {\em Right}: Duck (909 elements).
  %with 305 nodes, 72 ground nodes
  %Elements are colored by their index in the planned extrusion sequence. Purple elements are printed first, and red elements are printed last.
  }
  \label{fig:klein_bottle_teaser}\label{fig:duck}
\end{figure}

Spatial frame structures %, {\em i.e.} discrete structures consisting of linear bars, 
are used extensively in architecture to represent objects that cannot be easily captured by surfaces or volumetric solids ({\it e.g.} the Klein bottle in figure~\ref{fig:klein_bottle_teaser}).
In construction, these structures are useful due to their high strength-to-weight ratios~\cite{tam2018,huang2018iass}. % when optimized.
%Despite their simplicity in form, they can convey important topological and geometrical features  
%When used as load-bearing forms, frame structures can be optimized geometrically or topologically to achieve high strength-to-weight ratio~\cite{tam2018}\cite{huang2018iass}. 
%However, despite their ubiquitous appearance and importance in various design, engineering and scientific fields, an automatic fabrication strategy is still missing in the context of additive manufacturing, preventing a smooth transition from the design of frame structures to its materialization.
Most existing printing systems deploy a 2.5D strategy where melted materials are accumulated layer upon layer along a fixed direction. % printing % by a 3-axis gantry machine.
These systems are unable to print general 3D frame structures due to their inability to print in arbitrary directions.
% in planes 
% This layer-based printing logic, while sufficient for most of the volumetric shapes, is not appropriate for 3D discrete frame structures because of prohibitively long fabrication time and demands for temporal support structure.
Robot manipulators have proven to be viable alternatives for fabricating these structures due to their additional capabilities afforded by extra degrees-of-freedom (DOFs)~\cite{hack2014mesh,yu2016acadia,tam2018,piker2019continuous}. 
% sometimes called spatial 3D printing, 
%This technique involves extruding and solidifying thermoplastic along prescribed linear paths in space to form spatial meshes or lattices, taking advantage of the precision and flexibility offered by industrial robots as multi-axis machines.
However, robotic spatial extrusion has only been applied in limited capacities due to the planning challenges imposed when fabricating large, irregular structures.
The robot must respect both collision and kinematic {\em geometric} constraints present in manipulation tasks, and each partial-structure must respect {\em structural} constraints that ensure correct construction.
In extrusion planning, a {\em stiffness} constraint, which prevents significant structural deformation, is the primary structural constraint.
%n each partial structure
% Extrusion is representative of many construction tasks
Existing algorithms both require strong {\em human} guidance to solve these problems~\cite{Huang2018}
%in order to reduce the search space 
and lack completeness guarantees~\cite{wu2016printing,huang2016framefab}.
% In contrast, volumetric, continuum shapes have been under the spotlight of 3D printing technology's development since its early years, and advancement in the manufacturing process has partially fueled corresponding advancements in computational design \citeme{} and structural optimization of continuum shapes \citeme{}.

% because of the anisotropy structural behavior incurred by the material deposition
% process \cite{fang2017thesis} and prohibitively long fabrication time.

We present an algorithmic treatment of extrusion planning that focuses on its mathematical form, probabilistically complete algorithms, and algorithms that scale empirically.
In particular, we identify a dichotomy between satisfying geometric and structural constraints; stiffness most significantly impacts decisions at the {\em beginning} of construction while collisions most significantly limit actions towards the {\em end} of construction.
% Analysis due to the sheer number of collision constraints
% Forward distance from ground
% Start from the most constrained satisfying location
From this insight, we find that, in isolation, forward state-space search is most effective for stiffness constraints but backward state-space search is most effective for geometric constraints.
We provide algorithms that efficiently plan in the presence of both constraints by globally performing a greedy backward search, using forward reasoning to bias the search towards stiff structures.
%Existing research ... Algorithmic potential is not fully explored... The contributions of this paper are two-fold: (1) we introduce a new class of task and motion planning problems that are complex geometrically... and (2) we identify key properties of the planning problem and discuss several search strategies... 
%The heuristic explored in this work provide guidelines for future TAMP research that involve robotic interaction with objects that have reactions that can be predicted by computational methods.
The contributions of this paper are:
\begin{enumerate}
    \item A formalization of robotic spatial extrusion in the presence of stiffness and geometric constraints;
    \item Efficient and probabilistically complete forward and backward state-space search algorithms;
    \item Prioritization heuristics that guide both stiffness and geometric decision-making;
    \item An investigation of the failure cases of these methods;  
    \item Validation of our methods both on long-horizon simulated and real-world extrusion problems.
    %\item Identify interplay between stiffness and geometry
    %\item Lookahead
    %\item Regression algorithms
\end{enumerate}
% Section numbers

\section{Related Work} \label{sec:related}

%This work builds on prior work in spatial extrusion, automated assembly, and task and motion planning.

%\subsection{Robotic Spatial Extrusion} \label{sec:related-extrusion}
%Architectural research for more than a decade has demonstrated potential both in scale and in complexity. However, existing work takes an {\it ad hoc} approach to generating the extrusion sequence and plans the robot's motions by annotating the end effector's workspace path, leaving the kinematics and control to the industrial robot's controller~\citeme. \YJ{cite DMS2019 paper from Daniel Piker}
%\YJ{Graphics:}

Most existing work on extrusion planning only addresses planning for a free-flying end effector.
Wu {\it et al.} gave an algorithm for planning without stiffness constraints that considers a fixed discretization of end-effector orientations.
%for a 5-DOF free-flying end effector.
%free-flying vs free-floating vs freeform
It performs backward peeling~\cite{wu2016printing} and computes a partial-ordering of elements that respects collision constraints.
Then, it orders elements in a manner that preserves connectivity and the partial ordering.
However, this procedure is incomplete because it rigidly commits to a single partial ordering.
% Either type is readily solved on its own by polynomial-time algorithms
% Support constraints require that every edge must be connected to already printed parts
%that finds a partial ordering in a fashion similar to our regression algorithm (section~\ref{sec:regression}) when using the graph heuristic (section~\ref{sec:distance-heuristic}). 
% 4.2  Resolving the Collision Graph
% 5.1  Integrating Collision Constraints
%However, the scheduling part is not complete, and stiffness constraints are not considered. 
%Only the end effector's collision is considered. \YJ{more work on this}
% https://dl.acm.org/doi/10.1145/2897824.2925966
% The two stage algorithm is somewhat similar to using the {\em StiffPlan} heuristic
%To widen the applicability of robotic spatial extrusion, recent work formulates the extrusion problem as a sequence and motion planning problem. 
Huang {\it et al.} proposed a constrained graph decomposition algorithm to guide the extrusion sequence search \cite{huang2016framefab}; however, their algorithm is also incomplete.
%; however, their algorithm does not consider the robot's kinematics. % and thus does not guarantee manipulation feasibility.
% and Huang {\it et al.}
% http://papers.cumincad.org/data/works/att/acadia16_298.pdf
% http://web.mit.edu/yijiangh/www/papers/a224-huang.pdf
%In the context of 3D printing in bioengineering, 
Gelber {\it et al.} presented a complete forward search algorithm for a 3-axis printer that minimizes the deformation of a structure~\cite{Gelber_Hurst_Bhargava_2018}.
%However, their technique assumes a single orientation of the end effector.
% 3-axis printer, and thus only considers
%using an approximate stiffness computation.
% 3-axis printer 
%that uses a heuristic compliance check: deflection of the cantilevered beam
%to generate printing sequences that enable micro-scale spatial 3D printing on a purpose-built isomalt 3D printer. 
%NP-Complete proof
% The planners used by Wu and Huang used a forward assembly search, but in both cases, a  disassembly search would likely enable a simpler approach.
% The benefits of a forward search (easier  conceptualization, concurrent planning and printing)  outweighed the computational cost of computing these additional constraints
% https://arxiv.org/pdf/1801.00527.pdf
%Although their method lacks manipulation constraints, as they are working with a gantry system, this paper is one of the few prior works that investigates the impact of search control on planner performance... \YJ{more work on this}
Choreo is the first extrusion planning system using a robot manipulator~\cite{Huang2018}.
%considering both structural and manipulation constraints.
Choreo decomposes extrusion planning   into a {\em sequence planning} phase, where it plans each extrusion, and a {\em transit planning} phase, where it plans motions between each extrusion.
% Strictly lazy planning
Because of this strict hierarchy, Choreo is incomplete as it is unable to backtrack in the event that transit planning fails to find a motion plan.
% Sequence planning is framed as a Constraint Satisfaction Problem (CSP) where variables are time indices and values are elements.
% The CSP is solved using a forward search over assignments of each time index that backtracks in the event that no valid assignment exists.
Choreo performs a forward search during sequence planning, using constraint propagation to prune unsafe end-effector orientations. % with the partially-extruded structure.
%As a result, Choreo is similar to the progression search presented in section~\ref{sec:progression}.
% Progression or look ahead
To make sequence planning tractable, Choreo requires a user-generated
%decomposition that specifies a 
partial ordering on elements. % layers of elements along the global z axis.

Task and Motion Planning (TAMP) involves planning both the high-level objectives as well as the low-level robot motions required to complete a multi-step manipulation task~\cite{srivastava2014combined,Toussaint2017Multi-boundDomains,garrettIJRR2018}.
For extrusion planning, the high-level decisions are the extrusion sequence, and the low-level motions are the extrusion and transit trajectories of the robot.
A key challenge of extrusion planning when compared to typical TAMP problems is that its planning horizon is often substantially longer.
Solution to most TAMP benchmarks involves fewer than 50 high-level actions~\cite{lagriffoul2018platform}, while extrusion problems may require over 900 extrusions (figure~\ref{fig:duck}).
At the same time, extrusion planning is less general than TAMP in several ways: 1) there is a single goal state 2) the robot's configuration is the only continuous state variable 3) every solution is an alternating sequence of movements and extrusions of a known length.
%while in TAMP, there may be infinitely-many goal states.
%Elements are the only manipulable objects, and they either have not been extruded or are placed at a specified pose.
%Furthermore, 
%In contrast, many TAMP problems require moving objects placements that are not specified {\it a priori}.
%However, the high-level form of a TAMP solution is generally unknown and potentially arbitrarily long. 
% https://ieeexplore-ieee-org.libproxy.mit.edu/stamp/stamp.jsp?tp=&arnumber=8411475
Similar to how specializing to pick-and-place subclasses of TAMP enables the design of efficient algorithms~\cite{krontirisRSS2015,han2018complexity}, we take advantage of these restrictions and structural properties to develop efficient algorithms that scale to large problems.
Extrusion planning can framed as {\em Multi-Modal Motion Planning} (MMMP)~\cite{HauserLatombe,HauserIJRR11}, motion planning subject to a sequence of {\em mode} constraints $\sigma$ on the feasible configuration space of the robot ${\cal M}(\sigma) \subseteq {\cal Q}$.
Often times, ${\cal M}(\sigma)$ might is a lower-dimensional submanifold of an ambient space ${\cal Q}$.
%MMMP addresses planning for a continuous-time system subject to discretely-changing mode constraints. %imposed on the system. % discrete-time
%Two modes are {\em adjacent} if ${\cal M}(\sigma) \cap {\cal M}(\sigma') \neq \emptyset$.
A critical component of MMMP is identifying {\em transition configurations} $q \in {\cal T}(\sigma, \sigma') \subseteq ({\cal M}(\sigma) \cap {\cal M}(\sigma'))$ between modes $\sigma, \sigma'$, which allow for a discrete {\em mode switch} from ${\sigma \to \sigma'}$.
%Once a sequence of modes and a corresponding sequence transition configurations are found, planning can be reduced to a sequence of single-mode constrained motion planning problems~\cite{stilman2010global,berenson2011task,kingston2019exploring}. % each defined by the currently active mode.
%$[\sigma_1, ..., \sigma_n]$
%$[q_1, ..., q_{n-1}]$
%{\em mode families}
%In section~\ref{sec:mmmp}, we formalize extrusion planning as a MMMP problem.
Hauser and Ng-Thow-Hing provide an algorithm for MMMP that performs a forward state-space search through the space of modes~\cite{HauserIJRR11}.
They prove that their algorithm is {\em probabilistically complete}~\cite{kavraki1998analysis,Lavalle06}, namely that it will solve any {\em robustly feasible}~\cite{KFIJRR11} MMMP problem with probability one. % in a finite amount of time
%by sampling mode transitions.
However, their algorithm blindly explores the state-space, which is intractable for the problems we consider.

%As a result, it is equivalent to the \proc{Progression} algorithm (algorithm~\ref{alg:progression}) in section~\ref{sec:progression} if the priority function $k(\eta) = 0$ defined on search nodes is zero.
% probabilistically complete for
% Utility functions

% While extrusion planning is a subclass of MMMP and thus could be solved using Random-MMP~\cite{HauserIJRR11}, we take advantage of the structure of our subclass due to its known horizon, factoring, and physical properties to provide much more efficient algorithms.
% Uninformed/brute-force search
% Random-MMP simply randomly expands the tree
% No laziness

\section{Extrusion Sequencing}

%We begin with formulating the problem of robotic spatial extrusion, highlighting the planning variables and involved constraints. 

% By frame shapes we refer to spatial structures that consist of linear elements. %A frame structure $M$ can be described by its \textit{topology} and its \textit{geometry} which encodes the assignment of its vertices to points in 3D. %
% \YJ{these matrix representation is not necessary}
% A frame structure with connectivity graph $(V, E)$ can be described using two matrices: $X \in \R^{|V| \times 3}$ and $T \in \mathbb{Z}^{|E| \times 2}$, where each row of $X$ specifies the $(x, y, z)$ position of a vertex and each row of $T$ specifies the beam connecting two vertex indices. Robotic sequence extrusion requires us to search for a extrusion sequence to traverse all the elements, while prescribing the robot's configurations during and between extrusion processes.

We begin by formulating spatial extrusion planning in the {\em absence} of a robot.
%in order to highlight the structural constraints that are involved.
A {\em frame} structure is an {\em undirected geometric graph} $\langle N, E \rangle$ embedded within $\R^3$.
% space frame
Let the graph's vertices $N$ be called {\em nodes} and the graph's edges be called {\em elements} $E \subseteq N^2$ where $m = |E|$. 
Each node $n \in N$ is the connection point for one or more elements at position $p_n \in \R^3$. %\langle x_n, y_n, z_n \rangle
%Every element $\vec{e} = \langle n, n' \rangle \in E$ is an {\em unordered} pair of nodes $n, n' \in N^2$.
Each element $e = \{ n, n' \} \in E$ occupies a volume within $\R^3$ corresponding to a cylinder of revolution about the straight line segment ${p_{n} \to p_{n'}}$. % with radius $r$.
%We only consider frames with straight elements.
%where the center of the bottom face is $p_{n}$ and the center of the top face is $p_{n'}$
A subset of the nodes $G \subseteq N$ are rigidly fixed to {\em ground} and thus experience a reaction force.
Each element $e = \{ n, n' \}$ can either be extruded from ${n \to n'}$ or ${n' \to n}$.
Let {\em directed} element $\vec{e} = \langle n, n' \rangle$ denote extruding element element $e = \{ n, n' \}$ from $n \to n'$.
%in the reverse direction $n' \to n$. % permuted
We will use the set $P \subseteq E$ to refer to a set of printed elements, representing a partially-extruded structure.
%$N_P = G \cup \{n, n' \mid \langle n, n' \rangle \in P\} \subseteq N$f
Let $N_P = G \cup \{n, n' \mid \{n, n'\} \in P\} \subseteq N$ be the set of nodes spanned by ground nodes $G$ and elements $P$.
Extrusion planning requires first finding an {\em extrusion sequence}, an ordering of directed elements $\vec{\psi} = [\vec{e}_1, ..., \vec{e}_{m}]$.
We will use $\psi$ to denote the undirected version of $\vec{\psi}$.
Let $\vec{\psi}_{1:i} = [\vec{e}_1, ..., \vec{e}_{i}]$ give the first $i$ elements of $\vec{\psi}$ where $i \leq m$.
%such that each partially-extruded frame $\psi_{1:i} = P_i \in \powerset{E}$ satisfies a set of constraints
%, such as, a connectivity constraint that for each $\vec{e}_i = \langle n, n' \rangle$, $n \in N_{P_{i-1}}$.
%Nodal translational and rotational displacement tolerances
% Nodes $n \in N = \{1, 2, ...\}$

\subsection{Stiffness Constraint}

%Beside commonly involved constraints in the existing planning research, the \textit{stiffness constraint} evaluates the physical state of the object that the robot is operating on. 
%Compared to the common TAMP benchmarks, where the true physical state is usually approximated by a black-box simulation engine, the stiffness constraint relies on the finite element analysis of linear frame structures. 
%Later in this paper, we show that many insights can be gained by extracting useful information from the FEM analysis, which contributes to the design and comparison of different search strategies (section \ref{}). 

% Thus, to help clarifying our discussions, we present the explicit mathematical formulas for the elastic deformation analysis in following section.

The key structural invariant that must hold throughout the extrusion process is a \textit{stiffness constraint} requiring the maximal nodal deformation to be below a given tolerance.
Each element experiences a self-weight load due to gravity, which causes the structure to bend.
%that the maximal nodal deformation due to gravity for a partially-extruded structure $P$
% (or other constantly presented load) 
%is bounded by a predefined tolerance at each extrusion step. 
%In this paper, we assume elements are made up of Polylactic Acid (PLA) plastic material.
%
We approximate uniformly-distributed self-weight loads by applying half the load at each end of the element and using the fixed-end beam equation for moment approximation~\cite{gere1997mechanics}.
% Gere, J. M. and Timoshenko, S. P., 1997, Mechanics of Materials, PWS Publishing Company.
%
%In the extrusion planning problems considered in this paper, all frame elements are assumed to be made of same Polylactic Acid (PLA) plastic material and have the same solid circular cross section geometry with the diameter 1.75 mm. 
%PLA material properties can be found in the Appendix~\ref{sec:appendix}.
%The considered load case is the self-weight load in the direction of opposite to gravity. 
%The uniformly distributed load caused by self-weight is approximated by lumping the load to the two ends of the element, where the force component of the load is approximated by dividing gravity force by two. 
%All the grounded nodes $n \in G$ are assumed to have all six DOF fixed, which allows the structure to develop tensile and moment reactions at the fixities to allow partial structures that cantilever.
% under certain external loads and supports 
The deformation of all the nodes is calculated using finite element analysis of linear frame structures \cite{McGuire_Gallagher_Ziemian_1999}. For a 3D frame structure, each node has six degrees of freedom (DOF) $(u_x, u_y, u_z, \theta_x, \theta_y, \theta_z)$, which correspond to the translational and rotational nodal displacements in the global coordinate system. 
%In this section, we highlight the steps to compute the nodal displacements and reaction forces given an external load. \YJ{through which we can ...}
% Then, we describe the specific load case and support conditions considered in spatial extrusion planning problems.
% In the context of spatial extrusion, we only consider the force and moment caused by the self-weight under the gravitational force.
% ~\cite{bathe2006finite}
Using linear basis functions and the local-to-global frame transformation, we can derive the beam equation to link the nodal load to nodal displacement in the {\em global} coordinate system~\cite{McGuire_Gallagher_Ziemian_1999}:
$K_{e} 
\begin{pmatrix}
\u_{n}, \u_{n'}
\end{pmatrix}^T = \mathbf{f}_e
$.
% $K^{\cL}_{e} 
% \begin{pmatrix}
% \u^\cL_{n}, \u^\cL_{n'}
% \end{pmatrix}^T = \mathbf{f}^\cL_e
% %\label{eq:local_stiff}
% $
% for each element $e = \{ n, n'\}$ where  $\u^\cL_n = (u^\cL_n, v^\cL_n, w^\cL_n,\phi^\cL_{x, n}, \phi^\cL_{y, n}, \phi^\cL_{z, n})$ representing the element's the displacement of the end node $n$ in the element's {\em local} coordinate system. 
% The $12 \times 12$ matrix $K^\cL_{e}$ is the element stiffness matrix in the {\em local} coordinate system~\cite{McGuire_Gallagher_Ziemian_1999}. 
%which is dependent on the material's Young's modulus $E$, Poisson ratio $\mu$, moment of inertia $J_x, I_y, I_z$, cross section area $A$, and element length $L(e)$. 
%The explicit formula of $K^\cL_{e}$ can be found at section 4.5.5 of ``Matrix Structural Analysis'' \cite{McGuire_Gallagher_Ziemian_1999}.
% The lumped external load and moment acting on node $n, n'$ in the local coordinate system is $\mathbf{f}^\cL_e = (\mathbf{f}^\cL_{n}, \mathbf{f}^\cL_{n'})^T \in \R^{12}$. 
% To solve for the nodal displacement of the whole frame structure, we first transform each local force %eq.~\ref{eq:local_stiff} 
% to a common {\em global} coordinate system.
% The transformation for each element is computed by
% $K^\cG_{e} = R_e^T K^\cL_{e} R_e
% %\label{eq:local_global_e_stiff}
% $
% where $R_e$ is the block-diagonal local-to-global rotation matrix. 
% %
Then, by concatenating all nodal DOF into a vector $\u = (..., u_{x,n}, u_{y,n}, u_{z,n}, \theta_{x,n}, \theta_{y,n}, \theta_{z,n}, ...)$ for $n \in N$, the system stiffness matrix $K$ is assembled using:
\begin{equation}
K_{ij} = 
\bigg\{\begin{array}{lr}
\sum_{e \sim (i,j)} K_{e}(\textrm{e-dof}(i), \textrm{e-dof}(j)) &\textrm{if } i \sim j\\
0 &\textrm{otherwise}\end{array}
%\right
\label{eq:assemble_stiffness_matrix}
\end{equation}
where $i \sim j$ indicates that the nodal DOFs $i, j \in \{1, ..., 6|N| \}$ are connected by an element, $e \sim (i,j)$ indicates that element $e$ connects DOFs $i, j$, and $\textrm{e-dof}(i)$ gives the corresponding index of the DOF $i$ in the local element system.
% \CG{Is there a way of presenting this that is more consistent with my notation (e.g. doesn't use $i,j$)? Also, it's okay if that requires moving away from matrix notation}
%in eq.~\ref{eq:local_stiff} and \ref{eq:local_global_e_stiff}.
The support condition specifies a set of fixed nodal DOF indices $\{ s_1, \cdots, s_{6 |G|}\} \subset \{ 1, \cdots, 6 |N|\}$. The assembled system stiffness equation $K \mathbf{u} = \mathbf{F}$ is rearranged in the form:
\begin{equation}
\begin{pmatrix}
K_{ff} & K_{fs} \\ 
K_{sf} & K_{ss}
\end{pmatrix} 
\begin{pmatrix}
\u_{f}\\
\mathbf{0}
\end{pmatrix} = 
\begin{pmatrix}
\mathbf{F_{f}}\\
\mathbf{F_{s}}
\end{pmatrix} 
\label{eq:reorder_stiffness_matrix}
\end{equation}
The submatrix $K_{ff}$ is positive definite (PD) if 
%the frame does not contain mechanism ({\it i.e.} the structure can move without any external force exerted) and 
all elements are transitively connected to a ground node.
%or a ungrounded sub-structure. 
%the stiffness matrix is singular
Then, the nodal displacement under the structure's load can be obtained by solving the following sparse PD linear system:
%\begin{equation}
$K_{ff} \u_f = \mathbf{F}_f.$
\label{eq:solve_linear_system}
%\end{equation}
Let the procedure $\proc{Stiff}(G, P)$ test whether a partially-extruded structure $P$ with ground nodes $G$ satisfies the given maximum displacement tolerance. 

% and the internal reaction force $\mathbf{f}^L_{e}$ for element $e$ can be recovered by:
% %
% \begin{equation}
% \mathbf{f}^L_{e} = K^L_{e} R_e \u^G_e
% \label{eq:internel_force}
% \end{equation}

% \begin{align}
% &\begin{pmatrix}
% F^G_{e(0)x}, F^G_{e(0)y}, F^G_{e(0)z},
% F^G_{e(1)x}, F^G_{e(1)y}, F^G_{e(1)z}
% \end{pmatrix}^T = \notag\\
% &\begin{pmatrix}
% 0, 0, \frac{w_z L_e}{2},
% 0, 0, \frac{w_z L_e}{2}
% \end{pmatrix}^T \notag
% \end{align}

% \noindent
% where $e(0), e(1)$ is the two end points of the element $e$, $w_z = \rho A g_z$ with $\rho$ is the material's density, $A$ is the cross sectional area, $g_z$ is the gravitational acceleration in the global z axis, and $L_e$ is the length of the element $e$. The moment component of the self-weight load can be obtained by using the fixed-end beam equation \citeme{} a nd applying the local-to-global coordinate transformation:

% \begin{align}
% &\begin{pmatrix}
% M^G_{e(0)x}, M^G_{e(0)y}, M^G_{e(0)z},
% M^G_{e(1)x}, M^G_{e(1)y}, M^G_{e(1)z}
% \end{pmatrix}^T = \notag\\
% &\begin{pmatrix}
% 0, 0, \frac{w_z L_e}{2},
% 0, 0, \frac{w_z L_e}{2}
% \end{pmatrix}^T. \notag
% \end{align}

% \begin{itemize}
%     %\item Stability?
%     \item Stiffness: Let the procedure $\proc{Stiff}(G, P)$ test whether printed elements $P$ satisfies the maximum displacement tolerance given ground nodes $G$. If $\proc{connected}(G, P)$ does not hold, then the stiffness matrix is singular and $\proc{Stiff}(G, P)$ also does not hold. % $\neg \proc{connected}(G, P)$
% %.and cannot be Cholesky factored for the linear solve
% \end{itemize}

\begin{defn} \label{defn:valid}
    An extrusion sequence $\vec{\psi} = [\vec{e}_{1}, \vec{e}_{2}, ..., \vec{e}_{m}]$ is {\em valid} if $\{e \in \psi\} = E$ and $\forall i \in \{1, ..., m\}.\; \proc{Stiff}(G, \vec{\psi}_{1:i})$ and $n_i \in N_{\vec{\psi}_{1:i-1}}$ where $\vec{\psi}_i = \vec{e}_i = \langle n_i, n_i' \rangle$.
% \begin{itemize}
%     \item $\psi$ is an ordering of $E$ % $\{\psi\} = E$ % permutation
%     % https://www.encyclopediaofmath.org/index.php/Permutation_of_a_set
%     \item $\forall i \in \{1, ..., m\}.\; \proc{connected}(G, \psi_{1:i}), \proc{Stiff}(G, \psi_{1:i})$ 
% \end{itemize}
\end{defn}

\section{Robotic Extrusion}

We consider extrusion planning performed by a single articulated robot manipulator with $d$ DOFs. % performs each extrusion.
Let ${\cal Q} \subset \R^d$ be the bounded configuration space of the robot where $q \in {\cal Q}$ is a robot configuration.
%that encompasses fixed obstacles and joint limits
% $q_0 \in \R^d$ be the initial and goal robot configuration
The robot executes continuous trajectories $\tau: [0, 1] \to {\cal Q}$ where $\tau(\lambda) \in {\cal Q}$ is the robot's configuration at time $\lambda$ for $\lambda \in [0, 1]$.
% paths
The robot must adhere to its joint limits as well as avoid collisions with itself, the environment, and the currently printed elements.
Let $Q: P \to {\cal Q}$ be a function that maps a set of printed elements $P \subseteq E$ to the collision-free configuration space of the robot $Q(P) \subseteq {\cal Q}$.
% $Q: \powerset{E} \to \powerset{{\cal Q}}$
When no elements have been printed, $Q(\emptyset)$ is the collision-free configuration space of the robot when only considering environment collisions, self-collisions, and joint limits.
Each additionally printed element weakly decreases the collision-free configuration space,  {\it i.e.}
\begin{equation}
    P \subseteq P' \implies Q(P') \subseteq Q(P). \label{eqn:cspace}
    % Not \iff because an element $e \in P'$, $e \notin $
\end{equation}
% Submodularity-like property?
To ensure $\tau$ can be safely executed given printed elements $P$, $\forall \lambda \in [0, 1].\; \tau(\lambda) \in Q(P)$.
Finally, let $f_p(q) = x_p \in \R^3$ and $f_o(q) = x_o \in \SO{3}$ be the forward kinematic equations for the position and orientation of the end effector when the robot is at configuration $q$.
%\item Forward kinematics $f(q) = \langle x_p, x_o \rangle$
%\item End-effector pose $x = \langle x_p, x_o \rangle$ where $x_p$ is position and $x_o$ is orientation

\subsection{Extrusion}

\begin{figure}[htb]
 \includegraphics{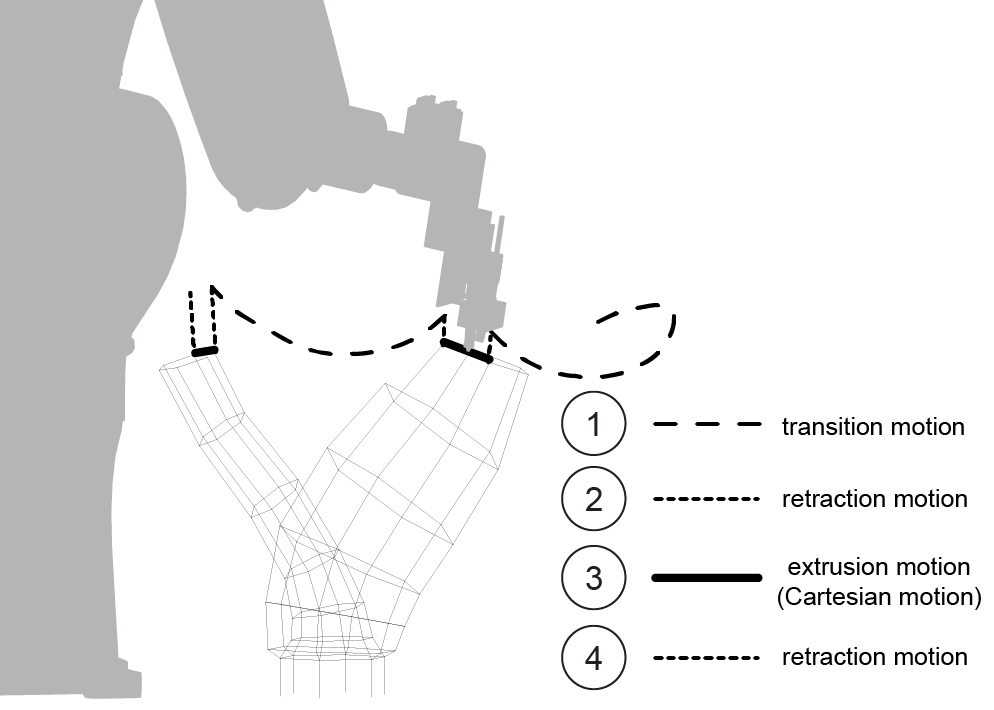}
 \caption{Transition, retraction, and extrusion motions for two elements.}
 \label{fig:hybrid_motion}
\end{figure}

The robot extrudes material at the position of its end effector while executing an {\em extrusion trajectory} $\tau_e$, which prints the continuous curve $l(\lambda) = f_p(\tau(\lambda))$.
% Line segment
% Extrusion trajectory $\tau_e \to {\cal Q}$ for element $e$ or $e^{-1}$.
Thus, element $\vec{e} = \langle n, n' \rangle$ can be extruded by following a trajectory $\tau_{\vec{e}}$ if $\forall \lambda \in [0,1]$:
% for $0 \geq \epsilon_p$
\begin{equation}
    ||\lambda p_{n} + (1 - \lambda) p_{n'} - f_p(\tau_e(\lambda))|| = 0. %\leq \epsilon_p
\end{equation}
%\item Monotonically increasing function $\lambda: [0, 1] \to [0, 1]$ such that $\lambda(0) = 0$ and $\lambda(1) = 1$
To prevent the end effector from colliding with the element while it is being extruded, the orientation of the end effector $x_o$ is constrained be within the hemisphere $X_o(\vec{e})$, the set of orientations opposite to the direction of ${p_{n} \rightarrow p_{n'}}$: % extrusion axes 
% Hemi-sphere constraint when printing
% $x_o \in X_o(n, n') \subset \SO{3}$ 
%, where $R(x_o)$ is the rotation matrix for orientation $x_o$.
\begin{equation*}
    % X_o(n, n') = \{x_o \in \SO{3} \mid (p_{n'} - p_{n})^\intercal R_{:,1}(x_o) \leq 0\}
    X_o(\langle n, n' \rangle) = \{x_o \in \SO{3} \mid \transpose{(p_{n'} - p_{n})} (x_o \cdot \transpose{[0, 0, 1]}) \leq 0\}.
    % \langle 1, 0, 0 \rangle
\end{equation*}
Additionally, we enforce that the end-effector orientation $x_o$ remains constant while extruding the element, $\forall \lambda \in [0, 1]$, $||x_o - f_o(\tau(\lambda))|| = 0$ to prevent the extruded material from inducing a twisting force.
% to make the fabrication possible. Otherwise the extruded material will induce some twisting force on the unsolidified element, which will make the extrusion process unstable/unpredictable.
%\begin{equation}
%    ||x_o - f_o(\tau(\lambda))|| = 0 % \leq \epsilon_o.
%\end{equation}
%Constant translational velocity and zero rotational velocity?
%Nonnegative extrusion end-effector translation $\epsilon_p$ and $\epsilon_o$ rotation tolerances
%Task-space constraints
In practice, we also require the robot to perform {\em retraction} motions that move into and out of contact with the extruded element without extruding any material.
Let $\rho \geq 0$ be an end-effector retraction distance hyperparameter.
Then, the retraction position for node $n$ at end-effector orientation $x_o$ is:
$r(n, x_o) = p_{n} + (x_o \cdot \transpose{[0, 0, -\rho]})$.
Thus, the end effector moves from ${r(n, x_o) \to p_n}$ before extruding $\vec{e}$ and from ${p_{n'} \to r(n', x_o)}$ after extruding $\vec{e}$.
We will treat retraction as a component of an extrusion motion.
% Contact with endpoints
%\addref \YJ{fig.~\ref{fig:hybrid_motion}}
See figure~\ref{fig:hybrid_motion} for a visualization of each motion type.

\subsection{MMMP Formulation} \label{sec:mmmp}

%The goal of this section is to make the jump from raw trajectories to mode sequences and motion planning problems
%As suggested in section~\ref{sec:tamp}, extrusion planning can be modeled as a MMMP problem. 
Viewing extrusion planning under this lens of MMMP is valuable for understanding the geometry of the problem and its impact on completeness.
Extrusion planning has two {\em mode families}, parameterized mode forms.
%We choose to exclude collision constraints from each mode in order to reason about motions that are safe with respect to one partially-extruded structure $P$ but potentially unsafe for another $P'$.
% Excluding collision constraints so we have factoring
% figure for the mode graph
A single {\em transit mode} (denoted as $\alpha$) governs the robot's movement while {\em not} extruding~\cite{AlamiTwoProbs,simeon2004manipulation}.
% $\sigma_t = \langle P \rangle \in \powerset{E}$
% transit path
% finite set of transit modes
%The only active constraints are collision constraints with printed elements $P \subseteq E$.
The only active constraint is trivially that $q \in {\cal Q}$.
Any probabilistically complete motion planner \proc{PlanMotion}, such as a Rapidly-Exploring Random Tree (RRT)~\cite{lavalle1998rapidly,8584061}, can be used to plan within transit modes.

% $\sigma_{\vec{e}} = \langle P, n, n', x_o \rangle \in (\powerset{E} \times N^2 \times \SO{3})$
An {\em extrusion mode} $\sigma_{\vec{e}} = x_o \in X_o(\vec{e})$ governs the robot's motion while extruding element $\vec{e} = \langle n, n' \rangle$ by starting at point $p_{n}$ and ending at $p_{n'}$. % with end-effector orientation $x_o$.
%, once gain, subject to collision constraints with printed elements $P \subset E$.
Here, $x_o$ is a continuous {\em coparameter} that defines the end-effector orientation constraint. % for mode $\sigma$.
% Can't extrude if P = E
Because of the position and orientation constraints on the end-effector, ${\cal M}(\sigma_{\vec{e}}) \subset {\cal Q}$ is a $(d - 5)$-dimensional submanifold of the ambient space ${\cal Q}$.
%$\Sigma = \Sigma_t \cup \Sigma_e$
As typical in constrained motion planning, we enforce that any trajectory $\tau_{}$ operating subject to mode $\sigma_{}$ stays within an $\epsilon$-neighborhood of ${\cal M}(\sigma_{})$~\cite{stilman2010global}. 
Let $\delta(q, {\cal M}(\sigma_{})) = \text{inf}_{q' \in {\cal M}(\sigma_{})} ||q - q'||$ be minimum distance from configuration $q$ to ${\cal M}(\sigma_{})$ and $\gamma(\tau, {\cal M}(\sigma_{})) = \text{sup}_{\lambda \in [0, 1]}  \delta(\tau(\lambda), {\cal M}(\sigma_{}))$ be the maximum distance from trajectory $\tau$ to ${\cal M}(\sigma_{})$.
%\begin{equation}
    % Apply left to right
    %g(\tau, {\cal M}(\sigma_{})) = \text{inf}_{q \in {\cal M}(\sigma_{})} \text{sup}_{\lambda \in [0, 1]}  || \tau(\lambda) - q||.
    %\forall \lambda \in [0, 1].\; \text{inf}_{q \in {\cal M}(\sigma_{})}  || \tau(\lambda) - q || < \epsilon.
    % If \leq \epsilon, can get nice open guarantees because trajectory cannot be touching constraint
    % This is violation rather than 
%\end{equation}
We enforce that the maximum constraint violation $\gamma(\tau, {\cal M}(\sigma_{}))$ is below a given $\epsilon > 0$.
% Make this epsilon clearance within the manifold
Any probabilistically complete single-mode {\em constrained motion planner}~\cite{stilman2010global,berenson2011task,kingston2019exploring} \proc{PlanConstrained} can be used to plan within extrusion modes.
% ${\cal T}(\alpha, \sigma_{\vec{e}})$ \neq \emptyset
Finally, let ${\cal T}(\alpha, \sigma_{\vec{e}}) = \{q \in {\cal Q} \mid f_p(q) = p_n, f_o(q) = x_o\}$ denote the set of {\em unidirectional} transition configurations from the transit mode to extrusion mode $\sigma_{\vec{e}}$, and ${\cal T}(\sigma_{\vec{e}}, \alpha) = \{q \in {\cal Q} \mid f_p(q) = p_{n'}, f_o(q) = x_o\}$ denote directed transition configurations from extrusion mode $\sigma_{\vec{e}}$ to the transit mode.

\subsection{Extrusion Problems}

\begin{defn}
    An {\em extrusion problem} $\Pi = \langle N, G, E, {\cal Q}, q_0 \rangle$ is defined by a set of nodes $N$, ground nodes $G$, elements $E$, configuration space ${\cal Q}$, and configuration $q_0 \in {\cal Q}$ specifying both the initial and final robot configuration.
\end{defn}
% Return to initial configuration or not?
% If so, can always check whether the initial configuration is reachable

\begin{defn} \label{defn:solution}
    For a given error threshold $\epsilon > 0$, a {\em solution} to an extrusion problem $\Pi$ is a valid extrusion sequence $\vec{\psi} = [\vec{e}_{1}, \vec{e}_{2}, ..., \vec{e}_{m}]$ (definition~\ref{defn:valid}), a sequence of extrusion mode coparameters $\vec{\sigma} = [\sigma_{\vec{e}_1}, ..., \sigma_{\vec{e}_m}]$, and an alternating sequence of $m + 1$ transit and $m$ extrusion trajectories $\pi = [\tau_{t_1}, \tau_{\vec{e}_1}, ..., \tau_{t_{m +1}}]$ such that:
    % for $\delta \geq 0$.
\begin{itemize}
    \item $\tau_{t_1}(0) = \tau_{t_{m+1}}(1) = q_0$
    \item $\forall i \in \{1, ..., m \}$.
    \begin{itemize}
        \item $\tau_{t_i}(1) = \tau_{\vec{e}_i}(0)$ % Endpoint?
        \item $\forall \lambda \in [0, 1].\; \tau_{t_i}(\lambda), \tau_{\vec{e}_i}(\lambda) \in Q(\psi_{1:i-1})$
        \item $\gamma(\tau_{\vec{e}_i}, {\cal M}(\sigma_{\vec{e}_i})) < \epsilon$
        %\item $\pi_{2i}(1) = \pi_{2i+1}(0)$ % Endpoint?
        %\item $\forall \lambda \in [0, 1].\; \pi_{2i}(\lambda), \pi_{2i+i}(\lambda) \in Q(\psi_{1:i-1})$
    \end{itemize}
    \item $\tau_{\vec{e}_m}(1) = \tau_{t_{m+1}}(0)$
    \item $\forall \lambda \in [0, 1].\; \tau_{t_{m+1}}(\lambda) \in Q(E)$.
    % \item $\forall j \in \{1, ..., |2E + 1|\}$ where $i = \lceil j/2 \rceil -1$.
    % \begin{itemize}
    %     \item $\pi_{j}(1) = \pi_{j+1}(0)$ % Endpoint?
    %     \item $\forall \lambda \in [0, 1].\; \pi_{j}(\lambda) \in Q(\psi_{1:i})$
    %     \item $\chi(\pi_{j}, \psi_{1:i}) \geq \delta$.
    % \end{itemize}
\end{itemize}
\end{defn}

\section{Algorithmic Tools}

% Fixed end-effector position trajectory. Maintain constant orientation
% Don't want he end-effector to ever move backward
% Illustrate the persistent component

We present state-space search algorithms for solving extrusion planning problems.
States $s = \langle P, q \rangle \in \powerset{E} \times {\cal Q}$ consist of the set of currently printed elements and the current robot configuration where $\powerset{E}$ denotes the power set of $E$.
The initial state is $s_0 = \langle \emptyset, q_0 \rangle$ and the goal state is $s_* = \langle E, q_0 \rangle$.
The \proc{Progression} algorithm (section~\ref{sec:regression}) performs a forward search from ${s_0 \to s_*}$, and the \proc{Regression} algorithm (section~\ref{sec:regression}) performs a backward search from the goal state ${s_* \to s_0}$.
%The {\em look ahead} algorithm (section~\ref{sec:lookahead}) augments the progression algorithm by enabling dead-end detection.
Both \proc{Progression} and \proc{Regression} perform a greedy best-first search~\cite{russell2016artificial} guided by a priority function $k(\eta)$ defined over {\em search nodes} $\eta$.
%  order-embedding
On each iteration, the search node $\eta$ in the {\em open list} $O$ that minimizes $k(\eta)$ is expanded. % priority queue $O$ 
% Total order
% https://en.wikipedia.org/wiki/Partially_ordered_set
% https://en.wikipedia.org/wiki/Best-first_search

The key trade off when designing these algorithms is the impact on satisfying stiffness and geometric constraints when searching forwards versus backwards.
For each constraint in isolation, it is advantageous to search from the {\em most constrained} state to the {\em least constrained} state.
At a less constrained state, the planner has more options and may prematurely make a decision that limits the legal options later in the search. %\citeme{}
In contrast, the forward or backward branching factor is generally small at the most constrained state, limiting the availability of poor choices.
Additionally, if the constrainedness either provably or empirically decreases over time, the pool of options will grow as the difficulty decreases.
%This often forces the 
%For feasible instances, the start and goal state guaranteed to be valid.
% Always start
Our algorithms leverage this principle, to search in directions that reduce the presence of {\em dead ends}, because in many extrusion problems, escaping dead ends can require an enormous amount of {\em backtracking} due to the long planning horizon.
%If a search is ever stuck in a {\em dead end}, it often must substantially {\em backtrack} in order to resume making progress.
%It is more efficient to start at the most constrained point in the problem, where the options are limited, and expand out like in RRT-Connect~\cite{KuffnerLaValle}.
We begin by developing common infrastructure for both the \proc{Progression} and \proc{Regression} algorithms.

\subsection{Sampling Extrusions}

%As stated in section~\ref{sec:mmmp}, we assume access to both \proc{PlanMotion} and \proc{PlanConstrained} procedures.
The key subroutine within each algorithm is \proc{SampleExtrusion} (algorithm~\ref{alg:extrusion}), which leverages \proc{PlanConstrained} to sample extrusion plans for an element $e$.
First, it samples a start node $n_1$ based on the currently printed nodes $N_P$.
This governs the extrusion direction $\vec{e} = \langle n_1, n_2 \rangle$.
Next, it samples an extrusion mode coparameter $\sigma_{\vec{e}} = x_o$ using \proc{SampleOrientation}.
% resulting in extrusion mode $\sigma = \langle P, n_1, n_2, x_o \rangle$.
This orientation produces the initial end-effector pose $\langle p_{n_1}, x_o \rangle$ and final end-effector pose $\langle p_{n_2}, x_o \rangle$.
Then, we use \proc{SampleIK}, an inverse kinematics procedure, to sample robot configurations $q_1, q_2$ that are kinematic solutions for these poses.
Finally, we call \proc{PlanConstrained} to find a trajectory from $q_1 \to q_2$ that satisfies mode constraints $\sigma_{\vec{e}}$ and does not collide with printed elements $P$.

%\CG{Lazy \proc{PlanMotion}}
%Lazy shortest path~\cite{bohlin2000path,dellin2016unifying}

\begin{algorithm}[hbt]
    \caption{Extrusion Sampling Algorithm}
    \label{alg:extrusion}
    \begin{algorithmic}[1] % The number tells where the line numbering should start
    \begin{small}
        \Procedure{SampleExtrusion}{$e, P; i$}
        \State $n_1 \gets \kw{sample}(\{n \in e \mid n \in N_P\})$
        \State $\{n, n'\} \gets e$
        \State $n_2 \gets n' \kw{ if } n_1 = n \kw{ else } n$
        \State $x_o \gets \proc{SampleOrientation}(n_1, n_2)$
        \State $q_1 \gets \proc{SampleIK}(p_{n_1}, x_o)$; $q_2 \gets \proc{SampleIK}(p_{n_2}, x_o)$ % Include i here?
        \State \Return $\proc{PlanConstrained}(q_1, q_2, x_o, P; i)$
        \EndProcedure
    \end{small}
    \end{algorithmic}
\end{algorithm}

\subsection{Deferred Evaluation} \label{sec:defer}

Standard state-space searches evaluate all feasible successor states $s' = \langle P \cup \{e\}, q' \rangle$ when expanding a state $s = \langle P, q \rangle$.
For extrusion planning, this requires planning both an extrusion trajectory $\tau_e$, where $q' = \tau_e(0)$, and a transit trajectory $\tau_t$ from ${q \to q'}$ for each remaining candidate element $e \in (E \setminus P)$. 
In the worst case, the number of successor ({\it i.e.} the branching factor) could be $\BigO{|E|}$.
This is exacerbated due to the fact that \proc{SampleExtrusion} and \proc{PlanMotion} are both computationally expensive due to collision-checking. % evaluating all successors can be 
%this wastes a substantial amount computational effort.
%However, assuming $s$ is not a dead end, only one successor state $s'$ will ultimately be included on a plan from $s$.
%As a result, the extrusion and transit trajectories for each other successor will not be used. 
To mitigate this problem, we adopt a {\em deferred evaluation}~\cite{helmert2006fast,richter2009preferred} strategy by planning extrusion and transit trajectories {\em after} popping a search node off the open list instead of {\em before} pushing the node on the open list.
%Delay the selection of the successor state
To enable this, search nodes in the open list are state and element pairs $\eta = \langle s, e \rangle$ where $e$ serves as ``action type'' that specifies the next element to be extruded. %, instead of simply a state $s$.
%Here, the element can be thought of as specifying the next {\em action} to try.
%Thus, full successor states $s'$ are {\em lazily} computed from a state and element pair $s, e$ where element $e$ intuitively serves as an action type. 
%\todo{} \YJ{I'm lost in this sentence, what do you mean by {\em action type}?}
% action mode
%In the event that no successors are feasible, the node is simply dropped.
This strategy dramatically reduces computation time, particularly in a greedy search, because it often avoids checking the feasibility of printing each successor element. %, where the discrete branching factor $b$ could be as large as $|E|$.
Once a feasible successor $s'$ is identified, the yet-to-be evaluated successors are deferred until the greedy search backtracks. 
% Branching factor $\BigO(|E|^2)$
% \BigOmega{|E|}
%We use a lazy best-first search~\cite{helmert2006fast,richter2009preferred} to order the open list.
% http://www2.informatik.uni-freiburg.de/~ki/papers/richter-helmert-icaps2009.pdf
% http://gki.informatik.uni-freiburg.de/papers/helmert-jair06.pdf
% http://www.fast-downward.org/Doc/SearchEngine

\subsection{Heuristic Tiebreakers} \label{sec:heuristic}
% https://en.wikipedia.org/wiki/Lexicographic_breadth-first_search
% TODO: likely shouldn't use $h(e)$
% \CG{Consider not using the name heuristic}

Because search nodes are state and element pairs, the priority function
$k(s, e)$ can take the next element $e$ into consideration.
We propose priority function $k(\langle P, q \rangle, e) = \langle r(P), h(e) \rangle$ that first orders search nodes by the number of {\em remaining elements} $r(P) = |E \setminus P|$ and lexicographically breaks ties using a {\em heuristic function} $h(e)$ defined on each individual element $e$.
By prioritizing search nodes where few elements remain to be planned, the search greedily explores the state-space in a {\em depth-first} manner.
% Perfect heuristic function
Because all successor states $s'$ of state $s$ have the same number of remaining elements $r$, the heuristic tiebreaker decides the order in which successors are considered.
This local ordering can have strong global effects on the sequence of partially-extruded structures considered.
% Stiffness and geometric
We consider four implementations of $h(e)$: (1) {\em Random}, (2) {\em EuclideanDist} and {\em GraphDist}, and (3) {\em StiffPlan}.

\subsubsection{Random Heuristic} \label{sec:random-heuristic}

The {\em Random} tiebreaker is a baseline where ties are broken arbitrarily. It orders elements by assigning each a value sampled uniformly at random $h(e) \sim U(0, 1)$. % from $[0, 1]$.

\subsubsection{Distance Heuristics} \label{sec:distance-heuristic}

The {\em EuclideanDist} and {\em GraphDist} heuristics prioritize elements that are close to ground, each according to a particular geodesic.
The {\em EuclideanDist} heuristic computes the Euclidean distance from the midpoint of element $e = \{n, n'\}$ to the ground plane.
When the ground plane is the xy-plane, this is simply the z-coordinate of the element's midpoint $h_e(e) = (p_n + p_{n'})/2 \cdot \transpose{[0, 0, 1]}$.
% Euclidean geodesic
% Distance in G
The {\em GraphDist} heuristic computes the minimum graph distance from any ground node $n \in G$ to the midpoint of element $e$ within the weighted frame geometric graph $\langle N, G \rangle$, where the weight of edge $e = \{n, n'\}$ is the Euclidean distance $||p_n - p_{n'}||$.
We precompute these distances upfront once by calling Dijkstra's algorithm starting from the set of ground nodes $G$.
% graph geodesic
% https://en.wikipedia.org/wiki/Geodesic
% https://en.wikipedia.org/wiki/Distance_(graph_theory)
%These distances are computed once upfront by solving for the shortest-path from a ground node to each node.
%Both of these heuristics prioritize elements that are close to ground in some space.
Intuitively, both of these heuristics guide the search through structures where the element load force has a short transfer path to ground because these structures are often stiff.
Additionally, these heuristics improve the sample complexity of \proc{SampleOrientation} because they often ensure end-effector orientations opposite to the z-axis remain feasible.
% angle from the z-axis
% Shortest paths from ground in some space

\subsubsection{Stiffness Heuristic} \label{sec:stiffness-heuristic}

The {\em StiffPlan} heuristic solves for a valid extrusion sequence $\vec{\psi}$, {\em ignoring} the robot, and uses the index $j$ of each element $e$ in the sequence ($\vec{\psi}[j] = e$) as its value $h_s(e) = j$. 
%total-ordering
%computes a valid extrusion sequence $\vec{\psi}$
Intuitively, because $\vec{\psi}$ is known to be stiff, it attempts to adhere to $\vec{\psi}$ as closely as possible subject to the additional robot constraints.
We compute a valid extrusion sequence $\vec{\psi}$ using a greedy forward search that is equivalent to \proc{Progression} in algorithm~\ref{alg:progression} if all robot planning is skipped. % finite version
We use the {\em EuclideanDist} heuristic $h_e$ (section~\ref{sec:distance-heuristic}) as the tiebreaker for this search.
%See the supplementary material 
See section~\ref{sec:plan-stiffness} for the full \proc{PlanStiffness} pseudocode. %\ref{sec:appendix}
% algorithm~\ref{alg:stiffness} in 
%The value of element $e$ is $h_s(e) = j$ for $\vec{\psi}[j] = e$, where $j$ is the index of $e$ in $\vec{\psi}$
%, ignoring the robot, and uses the index of each element $e \in \psi$ as its value. 

% Heuristics
% \begin{itemize}
%     \item Oppose gravity
%     \item Upfront once vs online
%     \item Stiffness vs distance vs collisions
%     \item Path distance heuristic (unweighted or weighted)
%     \item Most constrained stiffness-wise to least constrained stiffness-wise
%     \item Stiffness heuristics: node displacement, fixity forces, sum of forces
%     \item Geometric heuristics
% \end{itemize}

The {\em EuclideanDist}, {\em GraphDist}, and {\em StiffPlan} heuristics each perform a {\em forward} computation from ground to produce their values.
As we will see in section~\ref{sec:stiffness}, moving in a forward direction proves to advantageous for satisfying the stiffness constraint.
Finally, these heuristics can be seen as applying ``soft'' partial-ordering constraints that steer the search but do not limit completeness.
This is in contrast to the hard partial-ordering constraints in prior work~\cite{wu2016printing, huang2016framefab,Huang2018} (section \ref{sec:related}).

\subsection{Persistence} \label{sec:persistence}

The procedures \proc{SampleExtrusion} and \proc{PlanMotion} use sampling-based algorithms and thus are are unable to prove infeasibility.
%Sampling-based planning algorithms are generally only {\em semi-complete}, meaning that they only are guaranteed to return a solution if the problem is feasible.
%Thus, they are unable correctly report that there is no solution if the problem is infeasible.
% Non-deterministic TM or separate thread
As a result, both procedures must be reattempted indefinitely and with an increasing number of samples $i$. %timeout
%have a timeout hyperparameter $i$ that governs the number of samples consider on each attempt.
% Worse, could be arbitrarily long paths
In order to ensure that \proc{Progression} and \proc{Regression} are probabilistically complete, they both are {\em persistent}~\cite{GarrettIROS15} searches, meaning that they repeatedly expand each search node in a round-robin fashion until a plan is found.
% (section~\ref{sec:theory})
Let $i \geq 0$ denote the number of times a search node has been expanded.
We implement persistence by simply using the pair $\langle i, k(s, e) \rangle$ as the key for search nodes in the open list $O$. % priority queue
This ensures that the search node with the fewest attempts is always expanded first.
% Critical that finite horizon here
%prioritizes the number of expansions before the priority function $k(s, e)$.
% breadth-first manner
After a search node is expanded, it is re-added to the search queue $O$ with incremented priority $i + 1$.
As a result, this search node will not be re-expanded until all other nodes in $O$ have been expanded $i$ times. 

\section{Progression} \label{sec:progression}

\begin{algorithm}[!h]
    \caption{Progression Algorithm}
    \label{alg:progression}
    \begin{algorithmic}[1] % The number tells where the line numbering should start
    \begin{small}
        \Procedure{Progression}{$N, G, E, {\cal Q}, q_0; h$}
        \State $O = [\langle 0, \langle |E|, h(e) \rangle, \langle \emptyset, q_0 \rangle, e, [\;] \rangle \kw{ for } e \in E \kw{ if } e \cap G \neq \emptyset ]$ 
        %\Statex \;\;\;\;\;\;\;\;\;\;\;\;\;\;$\kw{ for } e \in E \kw{ if } e \cap G \neq \emptyset ]$
        %\While{$O \neq [\;]$}
        \While{\kw{True}}
            % Dynamic programming here instead?
            \State $i, \langle r, \_ \rangle, \langle P, q \rangle, e, \pi \gets \kw{pop}(O$)
            \State $P' \gets P \cup \{e\}$ % After addition
            \If{\kw{not} \proc{Stiff}$(G, P')$}
                \State \kw{continue} \Comment{No successors}
            \EndIf
            %\State $\pi' \gets \kw{None}$
            \State $\tau_e \gets \kw{None}$
            \If{$\proc{ForwardCheck}(E, G, P'; i)$} \Comment{Optional}
                \State $\tau_e \gets \proc{SampleExtrusion}(e, P; i$) \Comment{Extrusion} %{\color{blue}(Extrusion)}
            \EndIf
            \If{$\tau_e \neq$ \kw{None}}
                \State $\tau_t \gets \proc{PlanMotion}(q, \tau_e(0), P; i)$ \Comment{Transit}
                \If{$\tau_t \neq$ \kw{None}}
                    \State $\pi' \gets \pi + [\tau_t, \tau_e]$
                    \If{$P' = E$} \Comment{All printed}
                        \State $\tau_t \gets \proc{PlanMotion}(\tau_e(1), q_0, E; i)$        \If{$\tau_t \neq$ \kw{None}}
                            \State \Return $\pi' + [\tau_t]$ \Comment{Solution}
                        \EndIf
                    \EndIf
                    $s' \gets \langle P', \tau_e(1) \rangle$
                    \For{$e' \in (E \setminus P')$}
                        %\If{$e' \cap N_{P'} \neq \emptyset$} \Comment{Connected}
                        \State \kw{push}($O, \langle 0, \langle {r-1}, h(e') \rangle, s', e', \pi'\rangle$)
                        %\EndIf
                    \EndFor
                \EndIf
            \EndIf
            % Could just check $\proc{PlanMotion}(\tau_e(1), q_0, E; i)$
            % Unable to do cost-sensitive planning then
            %\If{$\pi' = \kw{None}$} % Can avoid readding if we assume connectivity
            \State \kw{push}($O, \langle i+1, \langle r, h(e) \rangle, \langle P, q \rangle, e, \pi \rangle$) \label{line:revisit} \Comment{Persistence}
            %\EndIf
        \EndWhile
        %\State \Return \kw{None}
        \EndProcedure
    \end{small}
    \end{algorithmic}
\end{algorithm}

% Shared meta-procedure

Algorithm~\ref{alg:progression} displays the pseudocode for \proc{Progression}.
%This algorithm is similar to the sequence planner used in Choreo~\cite{Huang2018} in that it is a greedy forward search.
%It performs a greedy forward state-space search.
Let $\pi$ be the currently planned trajectories for a search node.
After popping a state $\langle P, q \rangle$ and next element $e$ from the open list $O$, \proc{Progression} first checks whether the new structure $P' = P \cup \{e\}$ is stiff, taking advantage of the computational cheapness of \proc{Stiff}.
If not, the search node can be pruned altogether.
Otherwise, \proc{SampleExtrusion} samples an extrusion trajectory $\tau_e$ for element $e$.
The initial configuration $\tau_e(0)$ then becomes the goal for a transit motion that is found using \proc{PlanMotion}.
If $P' = E$, then the structure is fully printed, and all that remains is for the robot to return to $q_0$.
Otherwise, all remaining elements $e' \in (E \setminus P')$ are added to $O$ as successor search nodes.
Finally, search node $\langle P, q \rangle, e$ is re-added to $O$ with sampling timeout $i+1$ to be re-expanded in the future (section~\ref{sec:persistence}). % persistently
% Line numbers %\ref{line:revisit}
%In appendix~\ref{sec:theory}, 
%In the supplementary material, 
In theorem~\ref{thm:progression}, we prove \proc{Progression} is probabilistically complete.

\begin{figure}[!h]
\includegraphics[width=1.0\columnwidth]{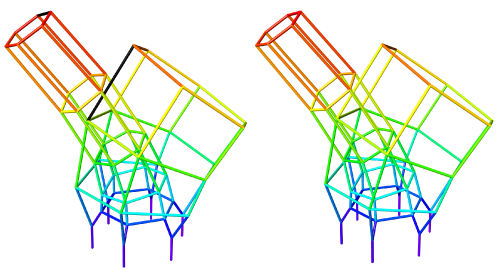}
 \caption{{\em Left}: The first state where \proc{Progression}-{\em EuclideanDist} backtracks (black elements are unprinted). {\em Right}: \proc{Regression}-{\em EuclideanDist} finds a solution without backtracking.}
 \label{fig:progression_failure}
\end{figure}

\proc{Progression} is geometrically sensitive to the extrusion sequence $\psi$. %the order that the elements are sequenced.
By equation~\ref{eqn:cspace}, when elements are added to $P = \{e \in \psi\}$, the collision-free configuration space $Q(P)$ weakly decreases.
As a result, \proc{SampleExtrusion} and \proc{PlanMotion} become more constrained as the plan grows.
In the worst case, $P$ may prevent some of the unprinted elements $E \setminus P$ from admitting any safe extrusions.
%extruding some elements can prevent later elements from being extrudable.
%For example, the search could extrude the exterior of a structure before the interior, forming a cage that prevents the robot from accessing the interior.
% Hollow shapes are easier
For example, figure~\ref{fig:progression_failure} demonstrates that \proc{Progression} becomes trapped in a dead end near the end of the horizon because it printed the left tail of the Klein bottle (figure~\ref{fig:klein_bottle_teaser}) before the black diagonal element.
In all of our structural figures, elements are colored by their index in a planned extrusion sequence. 
Purple elements are printed first, red elements are printed last, and black elements have yet to be printed.
%base substructure of the Klein bottle model shown in figure~\ref{fig:klein_bottle_teaser}.

% \begin{thm} 
%     Progression is probabilistically complete for robustly-feasible extrusion problems. \label{thm:progression}
% \end{thm}
% progression failure cases: geometric dead ends.

%Semantic attachments~\cite{dornhege09icaps}
%Impact of deferring motion planning for a progression
%All states reachable from start

%What happens when the start is the goal
%Connected components

\subsection{Forward Checking for Dead-End Detection} \label{sec:lookahead} % Lookahead
% https://en.wikipedia.org/wiki/Look-ahead_(backtracking)
% https://en.wikipedia.org/wiki/Local_consistency

In order to help \proc{Progression} avoid making poor geometric decisions, we developed a {\em forward-checking} (look ahead) algorithm~\cite{haralick1980increasing,dechter2003constraint} that is able to detect dead ends earlier in the search.
Intuitively, the robot must extrude every element in the structure eventually.
If there is ever an element that cannot be extruded given the partially-extruded structure $P$, then this state is a dead end.
Thus, \proc{ForwardCheck} eagerly evaluates the viability of many successors.
However, this acts oppositely to deferred evaluation (section~\ref{sec:defer}), and thus achieves better dead-end detection at the expense of worse computational overhead.
%As a result, \proc{ForwardCheck} is able better prune dead-end branches at the expense of worse computational overhead per expansion.
% Branching factor
% Geometric proximity
As a compromise, 
%Rather than sample an extrusion trajectory for all unprinted elements $E \setminus P$, 
we plan {\em extrusion trajectories} for only the elements $e$ that can {\em currently} can be printed given $P$, ({\it i.e.} $e \cap N_{P} \neq \emptyset$).
Intuitively, these elements are close in proximity to the printed structure and thus are most likely to be affected by a proposed geometric decision.
%Extruding elements that are not direct successors adds hefty overhead without significantly improving dead-end detection.

\begin{figure}[htb]
\includegraphics[width=1.0\columnwidth]{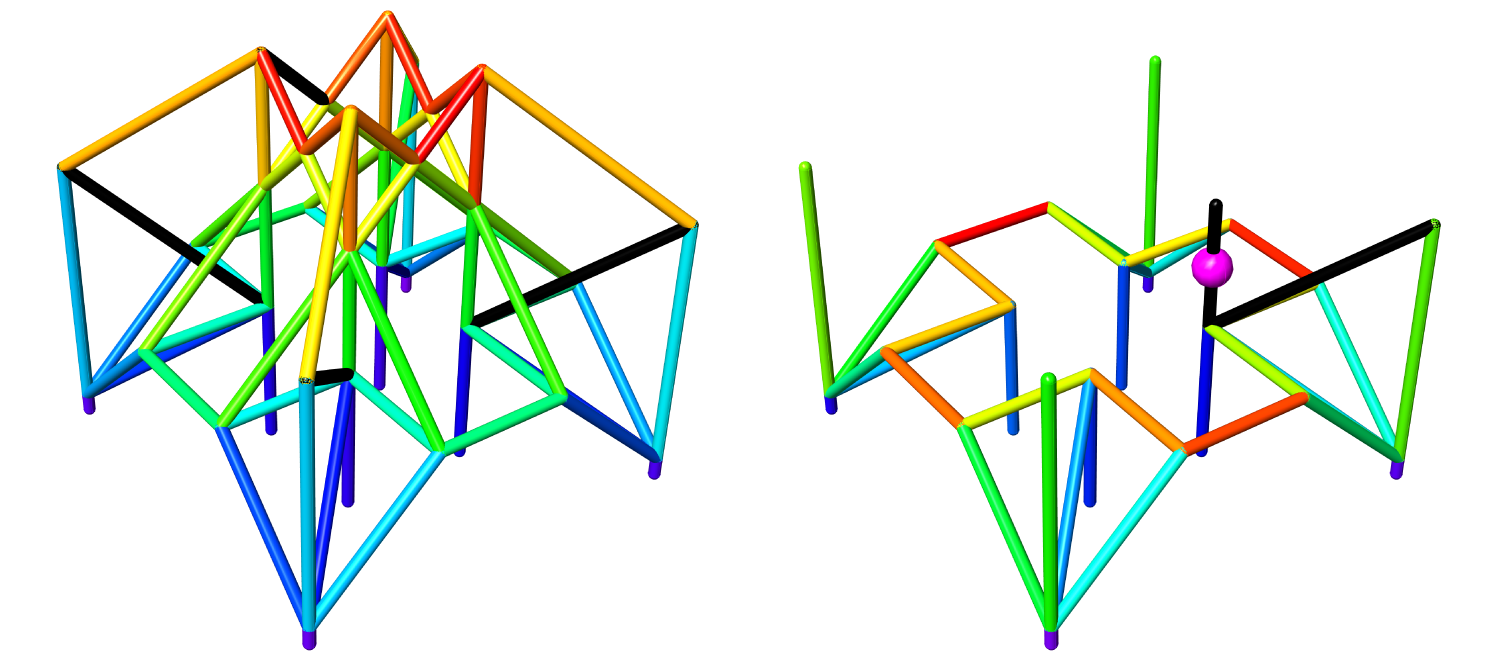}
 \caption{{\em Left}: the first state where \proc{Progression}-{\em GraphDist} backtracks (black elements are unprinted). {\em Right}: \proc{ForwardCheck} detects that printing the element indicated by the pink sphere prevents the diagonal black element from being safely extrudable.}
 \label{fig:lookahead_comparison}
\end{figure}

Algorithm~\ref{alg:lookahead} displays the pseudocode for \proc{ForwardCheck}.
%Let \proc{ForwardCheck} denote \proc{Progression} + \proc{ForwardCheck}.
% Use cache in the main search
It maintain a global {\em cache} of extrusion trajectories in order to reuse previously computed trajectories if possible.
%Printing for all elements results in many trajectories that never will be used.
%Additionally, some of the computed trajectories will not appear
Because \proc{ForwardCheck} invokes \proc{SampleExtrusion}, it cannot prove that a search node is a dead end. 
Thus, \proc{ForwardCheck} also uses the increasing sampling timeout $i$ to search for longer extrusion trajectories. 
%However, we develop a persistent (section~\ref{sec:persistence}) version of \proc{ForwardCheck} that continues to search for extrusions with increasing timeout $i$ until it succeeds.
%$\proc{ForwardCheck}(E, G, P; i) = \kw{True}$
%However, the success of \proc{ForwardCheck} can be used as a requirement before expanding the search node.
%Thus, we include a timeout for $i$ that grows as the search node is reattempted.
Figure~\ref{fig:lookahead_comparison} demonstrates an instance where \proc{ForwardCheck} detects, and thus avoids, a dead end early in the search.
The element with the pink sphere is the candidate element $e$ to be printed.
However, printing $e$ prevents the diagonal black element from being printable.
As a result, the search defers expanding $e$ at this time.

%Additionally, we {\em lazily} check collisions between each extrusion trajectory $\tau_e$ and partial structure $P$. \YJ{I think we need more explanation on {\em laziness}}.
%Thus, this avoids checking collisions for extrusion trajectory $\tau_e$ and element $e'$ when $e$ comes before $e'$. %$$e \prec e'$
% Just-in-time
% https://en.wikipedia.org/wiki/Principle_of_deferred_decision

% In contrast to Choreo~\cite{Huang2018}, we compute joint-space plans for each printable element instead of just end-effector plans.
% We found that using just end-effector plans resulted in poor dead-end detection, likely because many safe end-effector orientations do not admit kinematic solutions.
% Additionally, Choreo checks end-effector plans for all unprinted elements, instead of just those that are printable.

% \YJ{refer back to the topopt-101\_tiny case above, compare results. When no manual decomposition is used, Choreo takes 580s to solve this problem,while look ahead uses 545s (not that significant in this case...)}

%From an planning point of view, forward checking is equivalent to the admissible delete-relaxation heuristic $h_{max}$~\cite{BonetG99,bonet2001planning}, which when $h_{max}(s) = \infty$, detects a dead end.

% Allow choice of either direction
%Could check reachability $\proc{PlanMotion}(q_0, \tau_e(0)), \proc{PlanMotion}(\tau_e(0), q_1)$

% Pruning
% \begin{itemize}
%     \item Conditioning on previous choices (for certain branching factor)
%     \item Dominated trajectories
%    \item Tiebreaker using the number of collisions
% \end{itemize}

\begin{algorithm}[bt]
    \caption{Forward Checking Algorithm}
    \label{alg:lookahead}
    \begin{algorithmic}[1] % The number tells where the line numbering should start
        \begin{small}
        \Procedure{ForwardCheck}{$E, G, P; i$}
        \State cache $\gets \{e: [\;] \kw{ for } e \in E\}$ \Comment{Global cache}
        \For{$e \in (E \setminus P)$}
            \If{$e \cap N_{P} = \emptyset$} \Comment{Printable}
                \State \kw{continue}
            \EndIf
            \If{$\kw{any}(\proc{Safe}(\tau_e, P) \kw{ for } \tau_e \in \text{cache}[e])$}
                \State \kw{continue} \Comment{Reuse existing}
            \EndIf
            % Remove one element here?
            \State $\tau_e \gets \proc{SampleExtrusion}(e, P; i$) \Comment{Extrusion}
            \If{$\tau_e = \kw{None}$}
                \State \Return \kw{False}
            \EndIf
            \State $\text{cache}[e] \gets \text{cache}[e] + [\tau_e]$
        \EndFor
        \State \Return \kw{True}
        \EndProcedure
    \end{small}
    \end{algorithmic}
\end{algorithm}

\begin{figure}[htb]
\includegraphics[width=1.0\columnwidth]{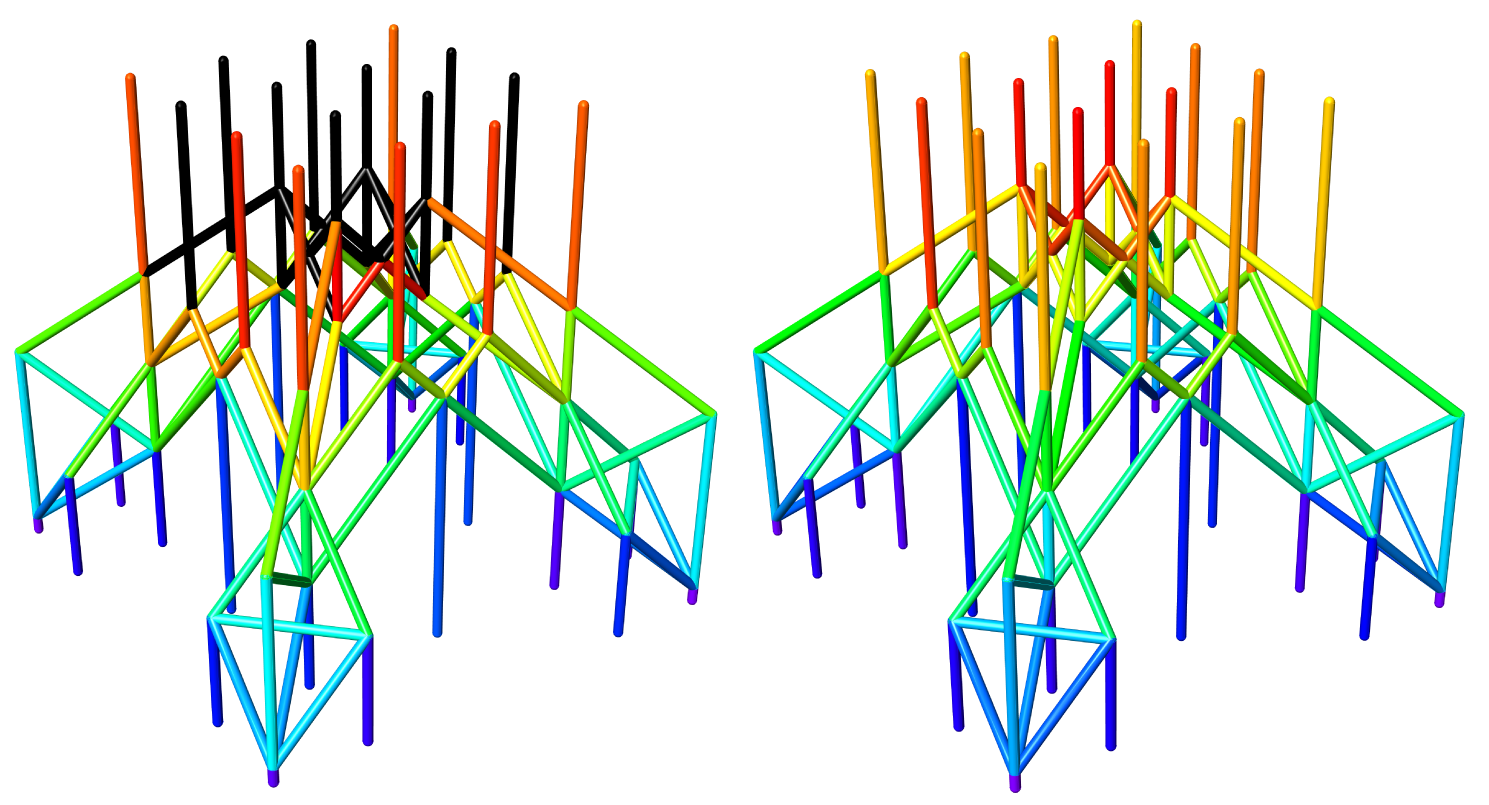}
 \caption{{\em Left}: the first state where \proc{ForwardCheck}-{\em GraphDist} backtracks (black elements are unprinted). {\em Right}: \proc{Regression}-{\em EuclideanDist} finds a solution without backtracking.}
 \label{fig:lookahead_failure}
\end{figure}

\proc{ForwardCheck} performs a one-step look ahead to detect dead ends.
However, it might the case that while each element can be printed individually, a {\em pair} of elements together cannot be printed.
If so, \proc{ForwardCheck} will not be able to detect the dead end until much later in the search, such shown in figure~\ref{fig:lookahead_failure}. 
Here, extruding any black element prevents at least one other nearby element from being safely printable. 
An {\em arc-consistency} look ahead that considers pairs~\cite{sabin1994contradicting} could detect these cases at the expense of even greater expansion overhead.
% maintaining arc-consistency (MAC) algorithm
% Failure cases: geometric dead ends caused by two elements conjunctively

\section{Regression} \label{sec:regression}

\proc{Regression} performs a backward search from the goal state to the initial state~\cite{nilsson2014principles,weld1994introduction,mcdermott1991regression,Ghallab04}.
In many planning domains, the goal conditions are under-specified, and as a result, there are many goal states.
Planners typically sample and plan from individual goal states; however, the set of goal states, and hence the initial branching factor, can be quite large.
Furthermore, sampled goal states might not be reachable from $s_0$, creating more opportunities for dead-end branches~\cite{bonet2001planning}.
% inconsistent, mutex
%or directly search over sets of states that all, upon applying a plan, satisfy the goal conditions.
Because extrusion planning has a single goal state $s_*$, these problems are avoided.
%the initial forward and backward branching factors are comparable,
%All states reachable from goal
% Highlight differences for regression
Algorithm~\ref{alg:regression} displays the pseudocode for \proc{Regression}.
The key differences from \proc{Progression} in algorithm~\ref{alg:progression} are that we negate $-h(e)$ in order to expand elements in the reverse order, the final extrusion configuration $\tau_e(1)$ is the start of each transit motion planning problem, and trajectories $[\tau_e, \tau_t]$ are prepended to plan $\pi$.
%In the supplementary material,
In theorem~\ref{thm:regression}, we prove \proc{Regression} is probabilistically complete.
% appendix~\ref{sec:theory}

\begin{algorithm}[bth]
    \caption{Regression Algorithm}
    \label{alg:regression}
    \begin{algorithmic}[1] % The number tells where the line numbering should start
        \begin{small}
        \Procedure{Regression}{$N, G, E, {\cal Q}, q_0; h$}
        \State $O = [\langle 0, \langle |E|, -h(e) \rangle, \langle E, q_0 \rangle, e, [\;] \rangle \kw{ for } e \in E]$
        %\While{$O \neq [\;]$}
        \While{\kw{True}}
            \State $i, \langle r, \_ \rangle, \langle P, q \rangle, e, \pi \gets \kw{pop}(O$)
            \State $P' \gets P \setminus \{e\}$
            % \If{\kw{not} (\proc{connected}$(G, P')$ \kw{and} \proc{Stiff}$(G, P')$)}
            %     \State \kw{continue} \Comment{No successors}
            % \EndIf
            \If{\kw{not} \proc{Stiff}$(G, P')$} % Returns false if disconnected
                \State \kw{continue} \Comment{No successors}
            \EndIf
            %\State $\pi' \gets \kw{None}$
            \State $\tau_e \gets \proc{SampleExtrusion}(e, P'; i)$ \Comment{Extrusion} %{\color{blue}(Extrusion)}
            \If{$\tau_e \neq$ \kw{None}}
                \State $\tau_t \gets \proc{PlanMotion}(\tau_e(1), q, P; i)$ \Comment{Transit}
                \If{$\tau_t \neq$ \kw{None}}
                    \State $\pi' \gets [\tau_e, \tau_t] + \pi$
                    \If{$P' = \emptyset$} \Comment{All printed}
                        \State $\tau_t \gets \proc{PlanMotion}(q_0, \tau_e(0); \emptyset, i)$
                        \If{$\tau_t \neq$ \kw{None}}
                            \State \Return $[\tau_t] + \pi'$ \Comment{Solution}
                        \EndIf
                    \EndIf
                    $s' \gets \langle P', \tau_e(0) \rangle$
                    \For{$e' \in P'$}
                        %\If{$e' \cap N_{P' \setminus \{e'\}} \neq \emptyset$} \Comment{Connected}
                        \State \kw{push}($O, \langle 0, \langle {r-1}, -h(e') \rangle, s', e', \pi'\rangle$)
                        %\EndIf
                    \EndFor
                \EndIf
            \EndIf
            %\If{$\pi' = \kw{None}$}
            \State \kw{push}($O, \langle i+1, \langle r, -h(e) \rangle, \langle P, q \rangle, e, \pi \rangle$) \Comment{Persistence}
            %\EndIf
        \EndWhile
        %\State \Return \kw{None}
        \EndProcedure
    \end{small}
    \end{algorithmic}
\end{algorithm}

% \begin{thm}
%     \proc{Regression} will solve any feasible geometry-only extrusion problem in polynomial time.
% \end{thm}

\begin{figure*}[!ht]
    \includegraphics[width=.25\textwidth]{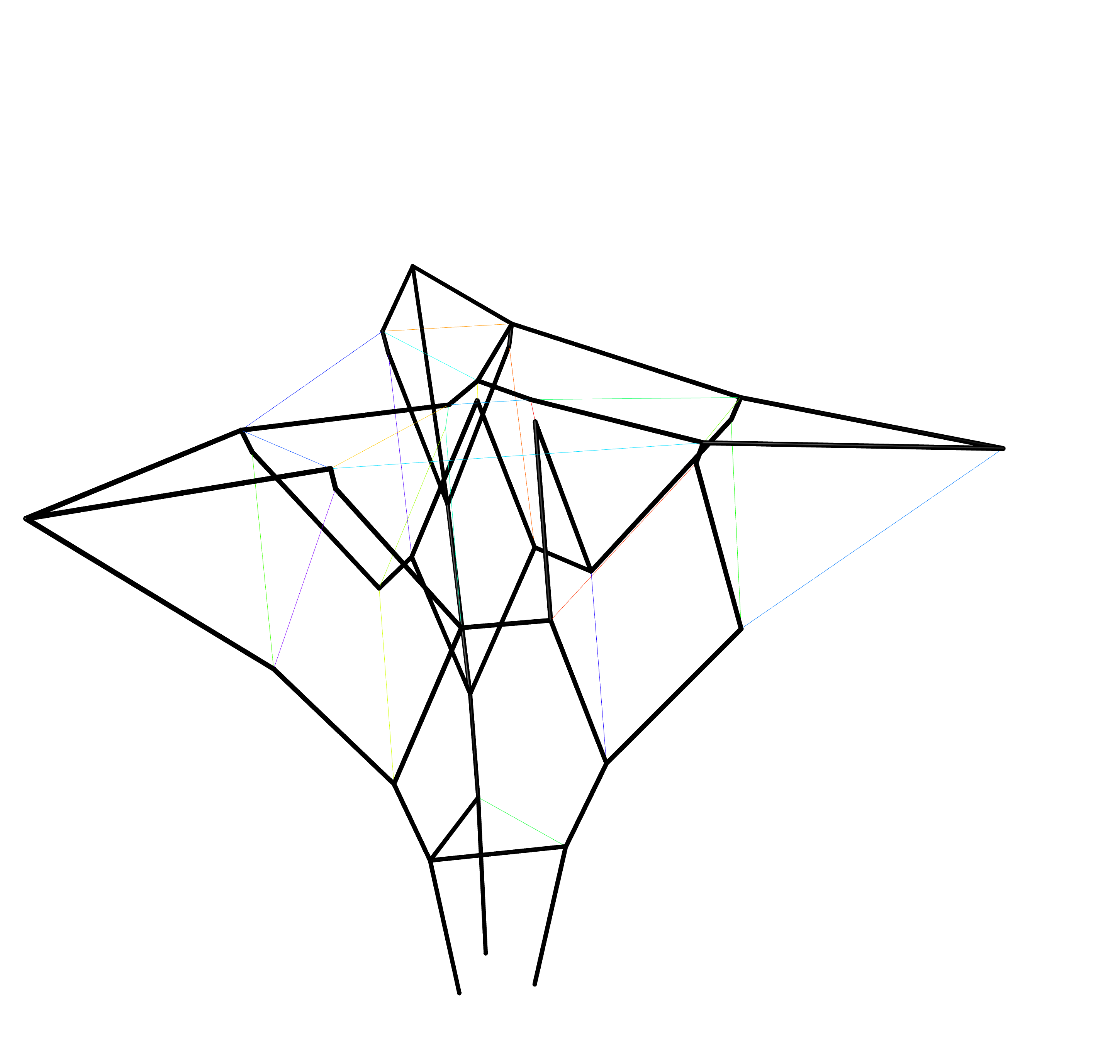}
    \includegraphics[width=.75\textwidth]{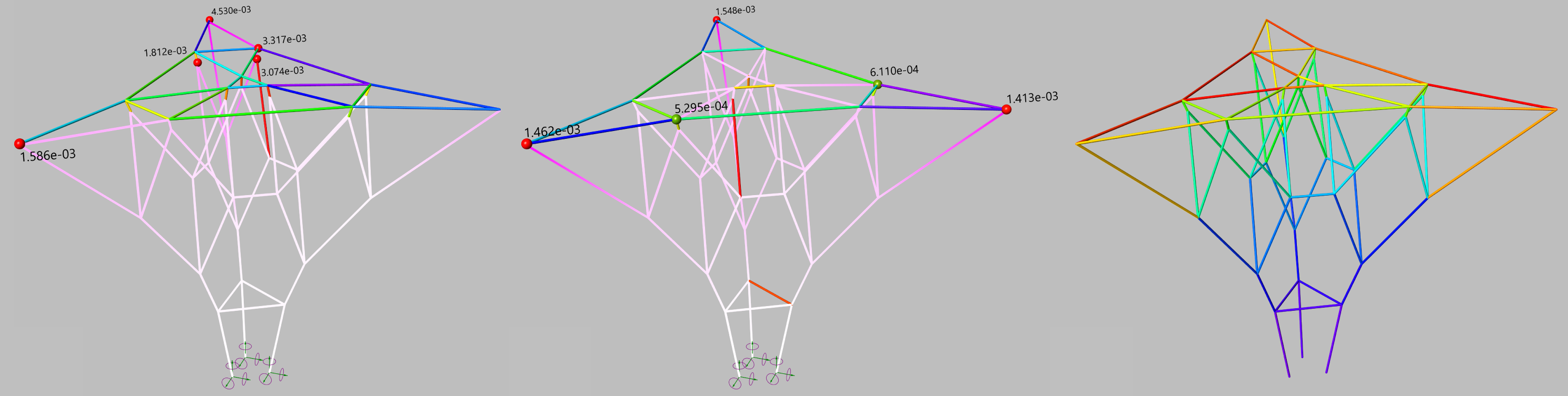}
 \caption{{\em From left to right}: 1) the unassigned substructure at the first state where \proc{Regression}-{\em Random} backtracks. 
 % and not stiff 
 %The black elements correspond to a substructure that both has not been sequenced and is {\em not} stiff. 
 2) the first state where \proc{Regression}-{\em EuclideanDist} backtracks. The element {\em deflection} is colored from white to pink. The five most deformed nodes are red and their translational displacements are annotated in meters % the norm of, the most deformed are 
 3) the first state where \proc{Regression}-{\em GraphDist} backtracks 
 4) \proc{Regression}-{\em StiffPlan} finds a solution without backtracking.}
 \label{fig:regression_failure}
\end{figure*}
% Oh in Fig.7, we need to explain the number's meaning: nodal translational displacement norm
% These nodes are also colored from green to red

\subsection{Geometric Constraints} \label{sec:geometry}

\proc{Regression} can be seen as {\em deconstructing} the structure by sequentially removing elements.
%as opposed to progression which builds the structure in a forward manner by adding elements.
From equation~\ref{eqn:cspace}, removing an element weakly increases the collision-free configuration space $Q(P)$.
%As a result, connectivity increases
Thus, the robot is the most geometrically constrained at the beginning of the search, limiting which elements can be initially extruded.
% Bug trap
% http://planning.cs.uiuc.edu/node219.html
% Branching factor more constrained at beginning than end
As a result, \proc{Regression}'s options with respect to geometry increase as the search advances, preventing it from being trapped in a geometric dead end.
%To motivate using a regression search to efficiently satisfy geometric constraints, we prove that \proc{Regression} has polynomial complexity when only considering collision constraints among a set of given extrusion trajectories.
% Infinite tolerance
To motivate using backward search to efficiently satisfy geometric constraints, we analyze a simplified {\em geometry-only} version of the extrusion problem that both omits stiffness and transit constraints as well as assumes a given set of possible extrusion trajectories $T$.
% stiffness-free
Given these simplifications, extrusion planning simply requires a identifying a totally-ordered subset of $T$ that extrudes each element exactly once.
% Could also do continuous version with sampling bound
% Constrained planning sampling rate
We consider a modified version of \proc{Regression} in algorithm~\ref{alg:regression} for extrusion-only problems.
Trivially, for all inputs, let $\proc{Stiff}(G, P) = \kw{True}$ and $\proc{PlanMotion}(q, q', P; i) = [q, q']$.
% Could you make any strange connectivity gadgets?
% Connectivity and collisions together could cause a circular dependency
% Directed trajectories
Additionally, 

\begin{small}
\begin{equation*}
    \proc{SampleExtrusion}(e, P; i) = \kw{sample}(\{\tau_e \in T \mid \proc{Safe}(\tau_e, P)\})
\end{equation*}
\end{small}
\noindent
arbitrarily selects a safe trajectory $\tau_e \in T$ for element $e$ if one exists.
Otherwise, \kw{sample} returns \kw{None}.
Under these conditions, it is easy to see that \proc{Regression} will solve feasible problem instances 
%without backtracking and 
in polynomial time (theorem~\ref{thm:complexity}).
%(see the supplementary material).

\subsection{Stiffness Constraints} \label{sec:stiffness}

% As demonstrated in figure~\ref{fig:stiffness_success}, forward search is able to more easily maintain the stiffness constraint than backward search.
% Intuitively, in the event that a partial structure $P$ is not stiff, the forward search will extrude other elements.
% %Often, there exists a set of elements that can support the problematic element $e$ 
% Often, the addition of extra elements $P'$ where $P \subset P'$ is able support $P$ by absorbing some of the load.
% As a result, the forward search is not susceptible to dead ends due to stiffness.
% %As a result, the forward search infrequently becomes stuck in dead ends due to stiffness. %that require a substantial amount of backtracking to escape.

Although \proc{Regression} makes geometric planning easier, it comes at the expense of increasing the difficulty of satisfying the stiffness constraint.
At the beginning of the backward search, there are many elements that can be removed without violating the stiffness constraint.
However, later in the backward search, there are fewer opportunities for supporting the structure, making the search more likely to arrive at a dead end caused by stiffness.
Figure~\ref{fig:regression_failure} image 1) shows the remaining-to-be-printed structure at the first dead end encountered by \proc{Regression}-{\em Stiffness}. 
As can be seen, arbitrarily removing elements sparcifies the structure and reduces its structural integrity.
To combat this, we use the heuristic tiebreakers in section~\ref{sec:heuristic} to bias the search to remain stiff.

To understand the impact of these tiebreakers on stiffness, we experimented on the extrusion problems in section~\ref{sec:results}, comparing the success rate of the \proc{Progression} and \proc{Regression} algorithms when {\em only} the stiffness constraint is active ({\it i.e.} ignoring the robot).
For \proc{Progression}, this is equivalent to \proc{PlanStiffness} in section~\ref{sec:stiffness-heuristic}. % and the supplementary material.
We performed 6 trials per algorithm, heuristic, and problem.
Each trial had a 5 minute timeout.
Figure~\ref{fig:all_success} displays the success rate of each algorithm.
\proc{Progression} was able to find an extrusion sequence for all problems, regardless of the heuristic. % that was used.
\proc{Regression} failed around 40\% of the time when randomly breaking ties.
However, \proc{Regression} was able to solve all problems when using the {\em StiffPlan} heuristic; although, this is not surprising given that {\em StiffPlan} explicitly uses a stiff plan. % to compute its values.
The {\em EuclideanDist} and {\em GraphDist} heuristics perform quite well but still have failure cases, such as in figure~\ref{fig:regression_failure}.
There, both heuristics prioritize removing the top of the structure, which is designed to provide tensile forces to hold the cantilevered elements~\cite{Lee_2018_phd}, causing the red vertices to deform significantly.

\section{Results} \label{sec:results}

\begin{figure*}[htb]
 \centering
 \includegraphics[width=0.32\textwidth]{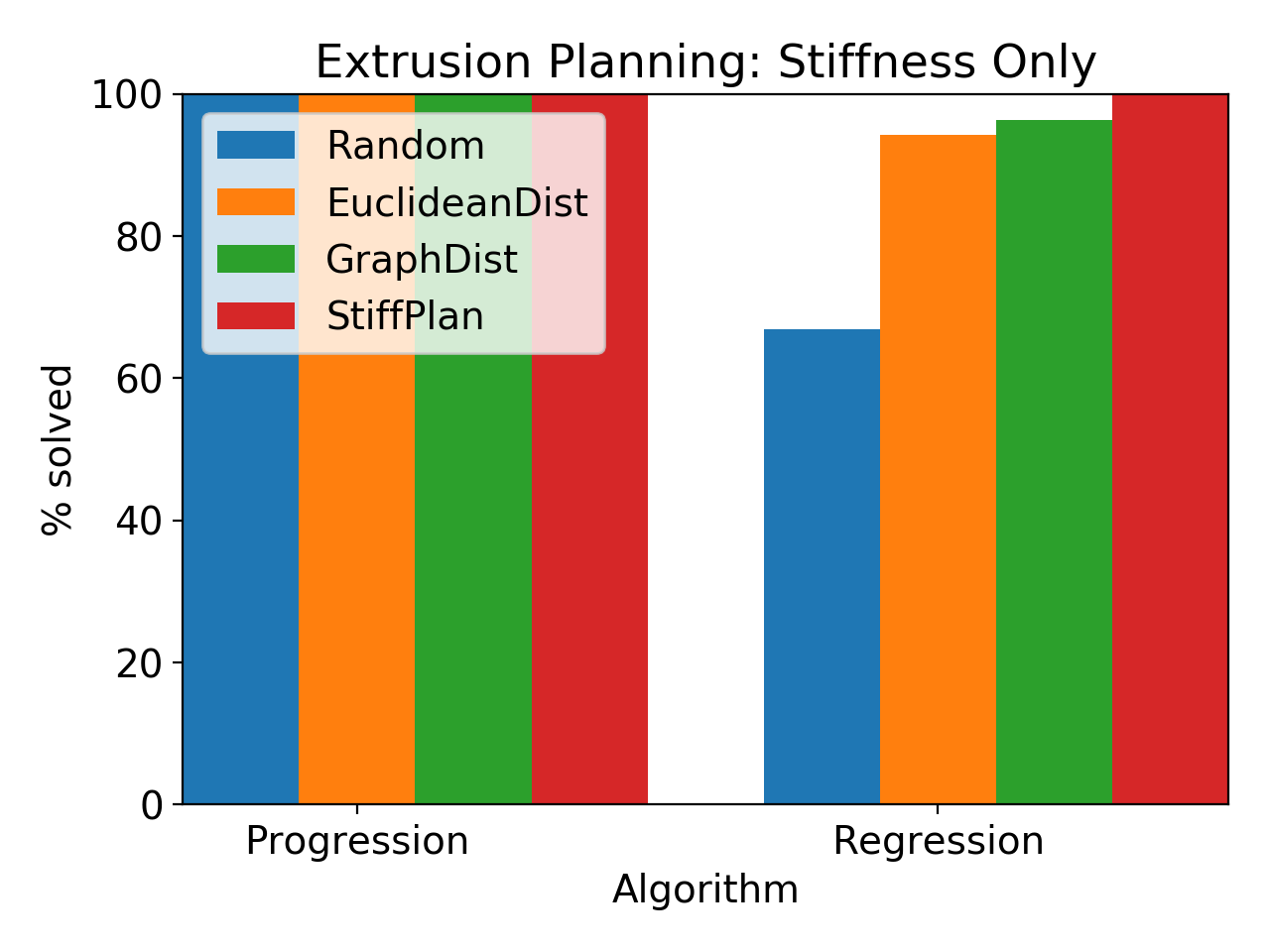}
 \includegraphics[width=0.32\textwidth]{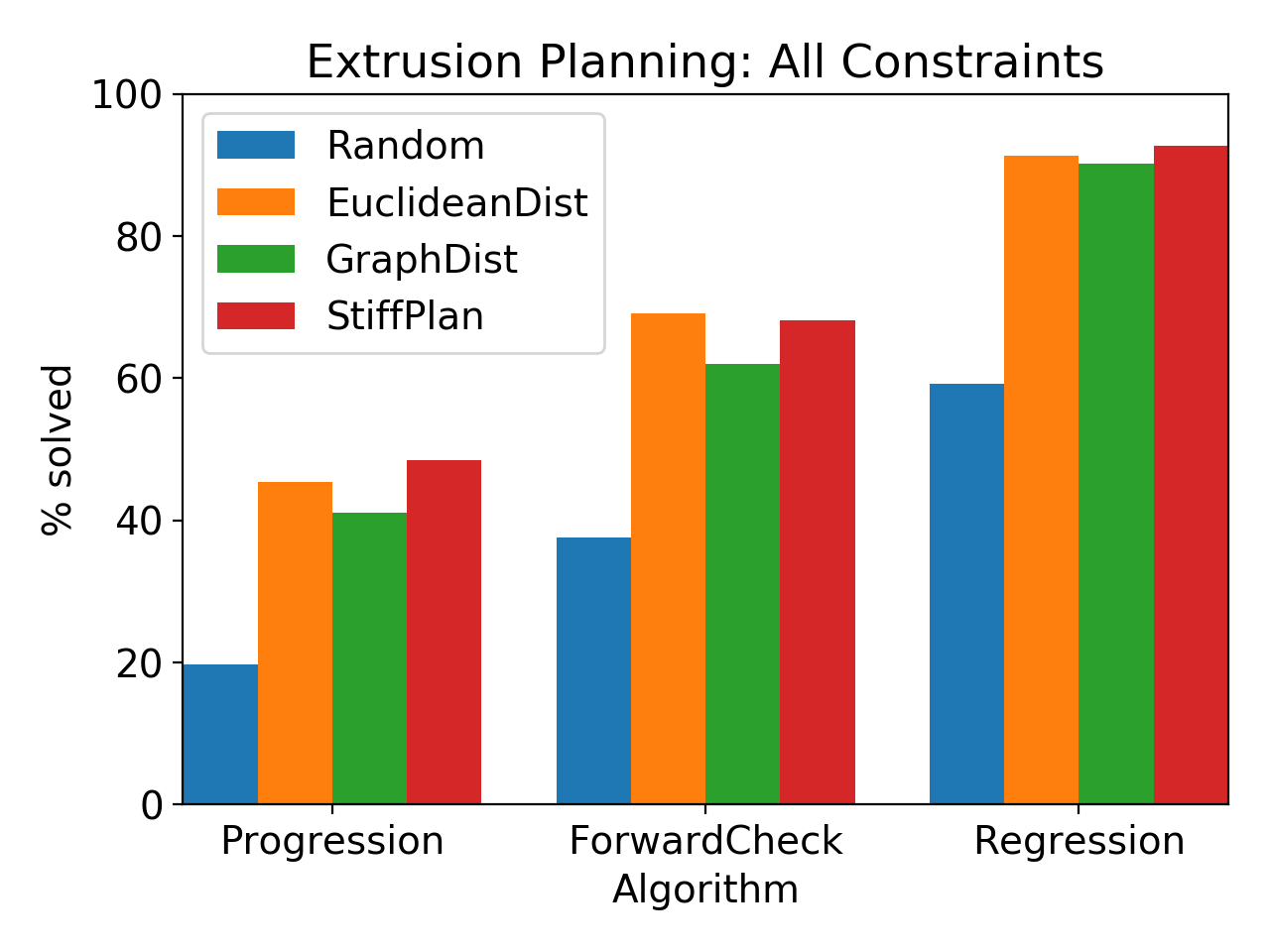}
 \includegraphics[width=0.32\textwidth]{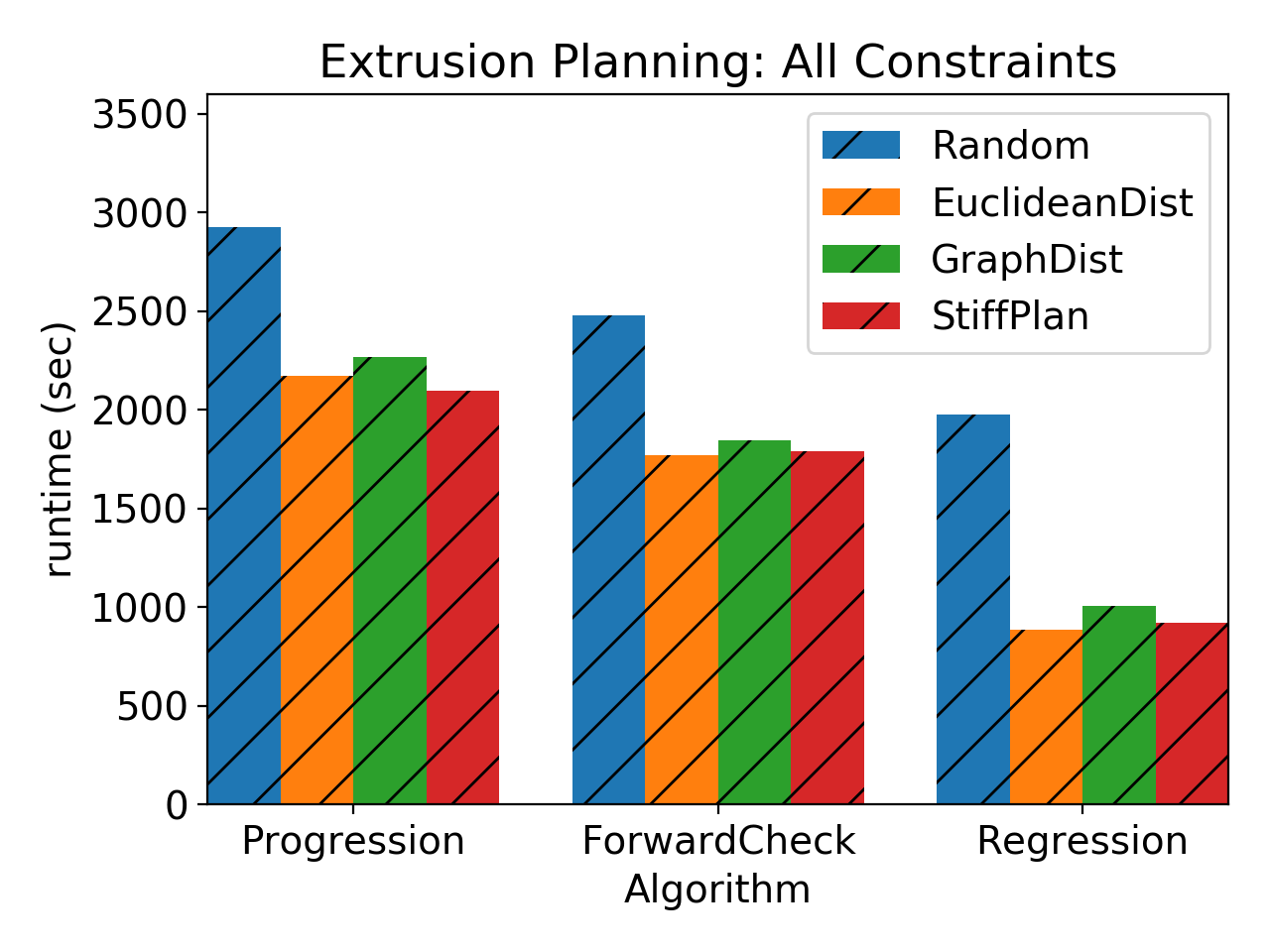}
 \caption{{\em Left}: the success rate of each algorithm (except \proc{ForwardCheck}) and heuristic pair subject to {\em only the stiffness constraint}. {\em Center}: the success rate of each algorithm and heuristic pair. {\em Right}: the average runtime in seconds of each algorithm and heuristic pair with a timeout of 1 hour (3600 seconds). 
 }
 \label{fig:all_success}
\end{figure*}
% python3 -m extrusion.analyze experiments/20-01-22_12-41-12.pk3 -a
% python3 -m extrusion.analyze experiments/20-01-22_15-22-34.pk3 -a

We experimented on 41 extrusion problems with up to 909 elements (the duck problem in figure~\ref{fig:duck}).
%See the supplementary material 
See section~\ref{appendix:benchmark} for a picture of each problem.
%For the camera-ready, we will open-source these problems as an extrusion-planning benchmark.
We experimented using all combinations of our 3 algorithms (\proc{Progression}, \proc{ForwardCheck}, and \proc{Regression}) and 4 heuristics ({\em Random}, {\em EuclideanDist}, {\em GraphDist}, and {\em StiffPlan}).
We performed 4 trials per algorithm, heuristic, and problem, each with a 1 hour timeout. % (3600 seconds).
%All constraints (stiffness, extrusion, transit, and collision) were enforced per trial.
%We implemented our planners in Python.
We used PyBullet~\cite{coumans2015bullet,coumans2019} for collision checking, forward kinematics, and rendering.
Because each element can only be in one pose, we preprocess the structure by computing a single, static axis-aligned bounding box (AABB) bounding volume hierarchy (BVH)~\cite{ericson2004real,kopta2012fast} for use during broadphase collision detection with each robot link.
% as well as the frame models 
We implemented \proc{PlanMotion} using RRT-Connect~\cite{KuffnerLaValle}, \proc{SampleIK} using IKFast, an analytical inverse kinematics solver~\cite{diankov2010automated}, and \proc{PlanConstrained} using Randomized Gradient Descent (RGD)\cite{yao2007path,stilman2010global} % lavalle1999probabilistic, yao2005path, stilman2007task
% an analytic inverse kinematics solver, when performing constrained motion planning.
% IKFast or Jacobian-based constrained planning
% Trajectory tracking
%For our KUKA KR6-R900 robot with $d = 6$ DOFs, IKFast analytically computes the full set of at most 16 inverse kinematics solutions.
%We will open-source our extrusion benchmarks and Python algorithm implementations.
See \url{https://github.com/caelan/pb-construction} for implementations of our algorithms.
%for the camera-ready copy of this paper. 
%\url{https://github.com/yijiangh/assembly_instances}

\begin{figure}[htb]
 \centering
 \includegraphics[width=1.0\columnwidth]{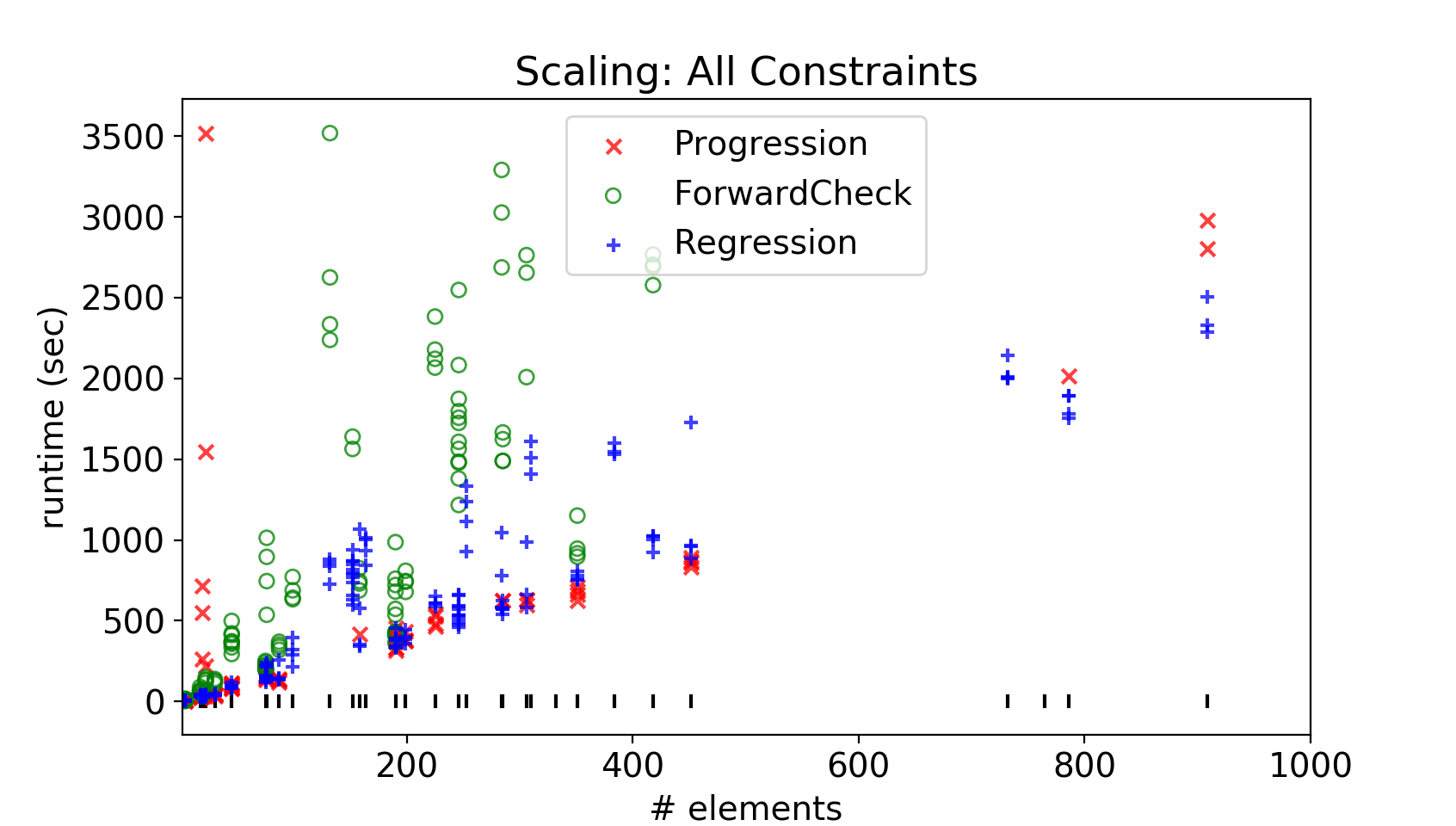}
 \caption{
    The runtime of each algorithm when using the {\em EuclideanDist} heuristic. The x-axis ticks denote the distribution of problem sizes.
   % The runtime of regression for each heuristic. Problem denotes the distribution of problem sizes.
 }
 \label{fig:scatter}
\end{figure}

Figure~\ref{fig:all_success} displays the success rate ({\em Center}) and the average runtime ({\em Right}) for each algorithm.
We assign a runtime of 1 hour for trials that failed to find a solution.
The {\em EuclideanDist}, {\em GraphDist}, and {\em StiffPlan} heuristics outperform {\em Random}, regardless of the algorithm.
The improved performance for both \proc{Progression} and \proc{Regression} indicates that the heuristics provide both stiffness and geometric guidance.
\proc{ForwardCheck} is able to solve more problems than \proc{Progression}, indicating that it is able to avoid some dead ends.
%some dead ends and thus outperform the uniformed progression search;
However, ultimately \proc{Regression} performed the best in terms of both success rate and runtime. 
%The results demonstrate that the regression algorithm performs best overall.
The best performing heuristic was {\em StiffPlan} followed closely by the {\em EuclideanDist}.
Our best-performing algorithms 
%(regression with either {\em EuclideanDist} or {\em StiffPlan}) 
are able to solve around 92\% of the problems and have an average runtime of about 15 minutes. % (900 seconds).
Figure~\ref{fig:scatter} ({\em Right}) displays the runtime of each trial per problem size when each algorithm uses the {\em EuclideanDist} heuristic.
Although \proc{ForwardCheck} is able to solve more problems than \proc{Progression}, it comes at the expense of longer runtimes.

% Can also remove the stiffness constraint
 
% \subsection{Adversarial}

% \begin{itemize}
%     \item Irregular designs
%     \item Rotations
%     \item Translations of structure
%     \item Different ground nodes
%     \item Scaling of structure: compas\_fea\_truss\_frame
% \end{itemize}

% Choreo's implementation is in C++ but Bullet is written in C++ and collision checking with Bullet is most expensive operation by far

We experimented on two extrusion problems considered by Choreo~\cite{Huang2018}. % in order to empirically compare our planners with prior work.
Choreo solves the ``3D Voronoi'' and ``Topopt beam (small)'' problems in 4025 and 3599 seconds whereas \proc{Regression}-{\em EuclideanDist} solves the problems in 742 and 2032 seconds.
Our planner outperforms Choreo despite the fact that Choreo had access to additional, human-specified information (section~\ref{sec:related}). 
%in the form of a human-specified decomposition (section~\ref{sec:related}).
% Their planner solves the {\em 3D Voronoi} in 4025 seconds while our best planner (\proc{Progression} + workspace) solves the problem in 620 seconds.
% Their planner solves the {\em Topopt beam (small)} in 3599 seconds while our best planner (regression + graph) solves the problem in 2212 seconds. 
% Choreo/PyChoreo (could rerun or compare with previous results)
%\subsection{Real-World Validation}
We validated our approach on three real-world extrusion problems. 
%See the supplementary material
See \url{https://youtu.be/RsBzc7bEdQg} for a video of our robot extruding each structure.
The largest of the three is the Klein bottle (figure~\ref{fig:klein_bottle_teaser}), which took about 10 minutes to plan for and 6 hours to print.
% klein_bottle_S1.5.json
% topopt-205_rotated_S1.5
% compas_fea_beam_tree_S_simp
%\ref{fig:regression_failure}
%a Klein bottle (figure~\ref{fig:klein_bottle_teaser}), a topology-optimized structure (TopOpt-205), and a tree structure (CompasTree, generated by \citeme).
% Klein bottle was about 6 hours
% Compas tree was within an hour
% Caelan R. Garrett*, Yijiang Huang*,
% Tomás Lozano-Pérez, & Caitlin T. Mueller 

% https://www.youtube.com/playlist?list=PLdlQ2M-oI1Dy9dFNnXbZXi2bWt-CP312b
% https://www.dropbox.com/sh/cng0fon58fraisp/AAByCKzPMFFo-XOQ2KPMfCTfa?dl=0

% \begin{itemize}
%     \item KUKA|PRC platform~\cite{wiki2018KRL}
%     \item Grasshopper~\cite{grasshopper2018} and Robots plugin~\cite{Robots2018}
%     \item Kangaroo~\cite{kangaroo2018}
%     \item Choreo/PyChoreo
%     \item Robot Operating System (ROS) Kinetic Release on Ubuntu 16.04~\cite{ROS}
%     \item ROS industrial~\cite{ROS-I2018Godel}
% \end{itemize}

\section{Conclusion}\label{sec:conclusion}

We investigated 3D extrusion planning using a robot manipulator.
Here, structural constraints are often at odds with geometric constraints.
Our algorithmic insight was to use backward search to plan geometrically feasible trajectories and to use forward reasoning as a heuristic that guides the search 
through structurally-sound states.
%towards stiff structures.
Future work involves extending our approach to general-purpose construction tasks.
%\cite{HoffmannN01}: Forward and Backward
% FF~\cite{HoffmannN01} is forward search with backward heuristic and you do the other.
% HSP~\cite{bonet2001planning} is backward search with forward heuristic

% \begin{itemize}
%     \item Current planning failure cases
%     \item Cost-sensitive planning
%     \item Additional construction domains
%     \item Multi-robot fabrication
%     \item Integrated design and planning (see Appendix \ref{sec:appendix})
% \end{itemize}

%\section*{Acknowledgments}

\newpage

% This year, we would like to use the ability of PDF viewers to interpret
% hyperlinks, specifically to allow each reference in the bibliography to be a
% link to an online version of the reference. 
% As an example, if you were to cite ``Passive Dynamic Walking''
% \cite{McGeer01041990}, the entry in the bibtex would read:

%% Use plainnat to work nicely with natbib. 

\bibliographystyle{plainnat}
\bibliography{references}

\newpage

\section{Stiffness Planning} \label{sec:plan-stiffness}

Algorithm~\ref{alg:stiffness} gives the pseudocode for \proc{PlanStiffness}, which implements the {\em StiffPlan} heuristic described in section V-C3.
It performs a greedy forward search similar to \proc{Progression} in algorithm 2, with the exception that the search is finite and does not involve the robot.
It uses the {\em EuclideanDist} heuristic $h_e$ (section~\ref{sec:distance-heuristic}) as its tiebreaker.
%(section~V-C2) 
\proc{PlanStiffness} is {\em complete} and will solve the extrusion sequencing problem in a finite (but not necessarily polynomial) amount of time.
In the event that \proc{PlanStiffness} returns \kw{None}, the extrusion planning problem is proved to be infeasible.

\begin{algorithm}[hbt]
    \caption{Stiffness Planning Algorithm}
    \label{alg:stiffness}
    \begin{algorithmic}[1] % The number tells where the line numbering should start
    \begin{small}
        \Procedure{PlanStiffness}{$N, G, E$}
        \State $O = [\langle \langle |E|, h_w(e) \rangle, \emptyset, e, [\;] \rangle \kw{ for } e \in E \kw{ if } e \cap G \neq \emptyset ]$ 
        \While{$O \neq [\;]$}
            % Dynamic programming here instead?
            \State $\langle r, \_ \rangle,  P, e, \vec{\psi} \gets \kw{pop}(O$)
            \State $P' \gets P \cup \{e\}$ % After addition
            \If{\kw{not} \proc{Stiff}$(G, P')$}
                \State \kw{continue} \Comment{No successors}
            \EndIf
            \State $\vec{\psi}' \gets \vec{\psi} + [e]$
            \If{$P' = E$}
                \State \Return $\{\vec{\psi}'[j]: j  \kw{ for } j \in \{1, ..., m\} \}$ \Comment{Solution}
                % \State \Return $\vec{\psi}'$ \Comment{Solution}
            \EndIf
            \For{$e' \in (E \setminus P')$}
                %\If{$e' \cap N_{P'} \neq \emptyset$} \Comment{Connected}
                \State \kw{push}($O, \langle \langle {r-1}, h_w(e') \rangle, P', e', \psi'\rangle$)
                %\EndIf
            \EndFor
        \EndWhile
        \State \Return \kw{None}
        \EndProcedure
    \end{small}
    \end{algorithmic}
\end{algorithm}
% I also run stiffness only versions of progression and regression

\section{Theoretical Results} \label{sec:appendix}

We state and prove the theoretical claims made in the paper.

\subsection{Regression Polynomial Complexity} \label{sec:complexity}

% NP-Complete when connectivity?

First, we analyze the complexity of \proc{Regression} for {\em geometry-only} extrusion problems (section~\ref{sec:geometry}).
%(section VII-A). 
Note that it is possible to achieve a better complexity of $\BigO{|T| |E|}$ using an algorithm that caches collisions.

\begin{thm} \label{thm:complexity}
    \proc{Regression} will solve any feasible geometry-only extrusion problem in polynomial time.
    %In the absence of stiffness and transit constraints and when given a fixed set of extrusion trajectories $T$, the complexity of \proc{regression} for {\em feasible} problems is $\BigO{|T| |E|}$.
    % Could include transit motions within this set
    % $\BigO{tm^2}$ if $t$ is the number of trajectories per element. $|T| = t|E|$
    % $\BigO{|T| m}^2$ if non collision caching
    \begin{proof}
        Each colliding pair $\neg \proc{Safe}(\tau_e, \{e'\})$ induces a partial-ordering constraint that element $e'$ must be extruded after element $e$ in order to safely execute trajectory $\tau_e$.
        By equation 3, %~\ref{eqn:cspace}, 
        removing element $e'$ weakly decreases the size of the set of partial-order constraints for each trajectory $\tau_e$.
        Because we assume feasibility, there exists a total ordering $\psi$ of $E$ and a corresponding sequence of trajectories $\pi$ from $T$ that respect collision constraints.
        As a result, for every set of unprinted elements $P' \subseteq E$, the element $e' = \psi[i] \in P'$ that has the largest index $i = \max_{\psi[j] \in P'}(j)$ in $\psi$ is guaranteed to have a safe trajectory $\tau_{e} \in T$.
        % Iterations are when successful
        Each of the $|E|$ iterations requires considering at most $|T|$ trajectories and checking collisions with at most $|E|$ elements.
        As a result, the complexity of \proc{Regression} is $\BigO{|T| |E|^2}$.
        %Assuming collision checks $\proc{Safe}(\tau_e, \{e'\})$ are cached, 
        %This process repeats for $|E|$ iterations.
        %$\BigO{|T| |E|}$ if we cache more effectively
    \end{proof}
\end{thm}

\subsection{Probabilistic Completeness} \label{sec:theory}

%Motion planning is PSPACE-Complete~\cite{canny1988complexity}. 
%TAMP is PSPACE-Complete % Vega-Brown
Because TAMP is decidable~\cite{deshpande2016tamp}, extrusion planning is also decidable, meaning that there exists {\em complete} algorithms that can correctly prove a problem is either feasible or infeasible.
%{\em semi-complete}, meaning that can only correctly prove a problem is feasible.
However, because we use randomized sampling-based strategies, we instead prove the weaker claim that our algorithms are probabilistically complete. % for robustly feasible extrusion problems. % in a finite amount of time with probability one.
% the weaker proposition
% Exponential convergence
First, we build on our problem formulation in section~\ref{sec:mmmp}.
%section IV-C.
by identifying a class of {\em robustly feasible}~\cite{KFIJRR11,garrettIJRR2018} extrusion problems, problems that admit a non-degenerate set of solutions making them amenable to sampling-based planning.
Define $\chi(\tau, P)$ to be the {\em clearance} of trajectory $\tau$~\cite{kavraki1998analysis} with respect to printed elements $P$ as the greatest lower bound on the distance from any configuration on $\tau$ to the boundary of the currently collision-free configuration space $\partial Q(P)$:
% to fixed obstacles, joint limits 
\begin{equation}
    \chi(\tau, P) = \text{inf}_{\lambda \in [0, 1]} \text{inf}_{q \in \partial Q(P)} || \tau(\lambda) - q ||.
\end{equation}
% clearance $\delta$
% Shortened elements
% Margin of distance
% Extrusion contact

% Sometimes it's feasible to assume there is a plan that exactly stays within the manifold. Other times it isn't
% Could apply $\epsilon$-neighborhood to the start and stop configurations which would ensure robust

%Let $\mu$ be a measure on $\SO{3}$. % $\mu:\SO{3} \to \R$
% https://en.wikipedia.org/wiki/3D_rotation_group
% https://en.wikipedia.org/wiki/Lie_group

% Epsilon feasibility in configuration space
% Measure in lower-dimensional spaces
% Measure transition configurations with respect to the manifold

\noindent
Let $\mu(X; {\cal X})$ be a measure on subsets $X \subseteq {\cal X}$ such that $0 < \mu({\cal X}; {\cal X}) < \infty$.
%$\tilde{\subseteq}, \subseteq_\emptyset$
Let $X \subseteq_\emptyset {\cal X} \implies [\emptyset \neq X \subseteq {\cal X}] \wedge [\mu(X; {\cal X}) > 0]$ denote that $X$ is a nonempty subset of ${\cal X}$ with positive measure with respect to ${\cal X}$.

% Apply the solution definition conditioned on a set of values
\begin{defn} \label{defn:robust}
    An extrusion problem $\Pi = \langle N, G, E, {\cal Q}, q_0 \rangle$ is {\em robustly feasible} for a valid extrusion sequence $\vec{\psi} = [\vec{e}_{1}, \vec{e}_{2}, ..., \vec{e}_{m}]$ (definition~\ref{defn:valid}) if there exists sequence of extrusion mode coparameter sets $[\Sigma_{\vec{e}_1}, ..., \Sigma_{\vec{e}_m}]$ s.t.
    % (section III-A)
    \begin{equation}
        \forall {i \in \{1, ..., m \}}. \Sigma_{\vec{e}_i} \subseteq_\emptyset X_o(\vec{e}_i) \label{eqn:mode}
    \end{equation}
    and $\forall \vec{\sigma} = [\sigma_{\vec{e}_1}, ..., \sigma_{\vec{e}_m}] \in \bigotimes_{i=1}^m \Sigma_{\vec{e}_i}.$ exists:
    \begin{itemize}
        \item a sequence {\em start} and {\em end} extrusion configuration sets $[T_{\sigma_{\vec{e}_1}}, ..., T_{\sigma_{\vec{e}_m}}]$ and $[T_{\sigma_{\vec{e}_1}}', ..., T_{\sigma_{\vec{e}_m}}']$ s.t.
        %\raggedright
        \begin{flalign}
            \forall i \in \{1, ..., m \}.& T_{\sigma_{\vec{e}_i}} \subseteq_\emptyset {\cal T}(\alpha, \sigma_{\vec{e}_i}) \label{eqn:start} \\
            & T_{\sigma_{\vec{e}_i}}' \subseteq_\emptyset {\cal T}(\sigma_{\vec{e}_i}, \alpha) \label{eqn:end}
        \end{flalign}
        and $\forall [q_{\sigma_{\vec{e}_1}}, ..., q_{\sigma_{\vec{e}_m}}] \in \bigotimes_{i=1}^m T_{\sigma_{\vec{e}_i}}$ \\and  $\forall [q_{\sigma_{\vec{e}_1}}', ..., q_{\sigma_{\vec{e}_m}}'] \in \bigotimes_{i=1}^m T_{\sigma_{\vec{e}_i}}'.$ exists:
        \begin{itemize}
            % (section IV-C)
            \item a {\em solution} (definition~\ref{defn:solution}) comprised of $2m+1$ trajectories $\pi = [\tau_{t_1}, \tau_{\vec{e}_1}, ..., \tau_{t_{m +1}}]$ s.t.
            \begin{flalign}
                \forall i \in \{1, ..., &m \}\;. \tau_{\vec{e}_i}(0) = q_{\sigma_{\vec{e}_i}}, \tau_{\vec{e}_i}(1) = q_{\sigma_{\vec{e}_i}}' \label{eqn:endpoints} \\
                &\chi(\tau_{t_i}, \psi_{1:i-1}), \chi(\tau_{\vec{e}_i}, \psi_{1:i-1}) > 0 \label{eqn:clearance}
            \end{flalign}
            and $\chi(\tau_{t_{m+1}}, E) > 0$.
        \end{itemize}
    \end{itemize}
\end{defn}
% Constrained motion planning literature
% Can also just test if robust trajectory back to $q_0$
%Let $\Sigma \subseteq \SO{3}$ be an open set of end-effector orientations.

Breaking down the definition, equation~\ref{eqn:mode} requires the mode set $\Sigma_{\vec{e}_i}$ for each extrusion to have positive measure with respect to the mode space for $\vec{e}_i$.
Equations~\ref{eqn:start} and~\ref{eqn:end} states that for each mode $\sigma_{\vec{e}_i} \in \Sigma_{\vec{e}_i}$, the set of transition configurations $T_{\sigma_{\vec{e}_i}}$ from ${\alpha \to \sigma_{\vec{e}_i}}$ and the set of transition configurations $T_{\sigma_{\vec{e}_i}}'$ from ${\sigma_{\vec{e}_i} \to \alpha}$ both have positive measure relative to their respective spaces.
Finally, equation~\ref{eqn:clearance} states that there exists solutions $\pi$ where the transit trajectory $\tau_{t_i}$ between the pair of transition configurations ${q_{\sigma_{\vec{e}_{i-1}}}', q_{\sigma_{\vec{e}_i}}}$ for transit mode $\alpha$ has positive clearance and the extrusion trajectory $\tau_{\vec{e}}$ between each pair of transition configurations ${q_{\sigma_{\vec{e}_{i}}}, q_{\sigma_{\vec{e}_i}}'}$ for extrusion mode $\sigma_{\vec{e}_i}$ has positive clearance.
%\ref{eqn:endpoints}
As a result, the motion planning problem ${q_{\sigma_{\vec{e}_{i-1}}}' \to q_{\sigma_{\vec{e}_i}}}$ and the constrained motion planning problem ${q_{\sigma_{\vec{e}_{i}}} \to q_{\sigma_{\vec{e}_i}}'}$ subject to manifold ${\cal M}(\sigma_{\vec{e}_i})$ are both robustly feasible.

We assume that \proc{PlanMotion} is a probabilistically complete motion planner and \proc{PlanConstrained} is a problematically complete constrained motion planner.
Assume that $\proc{SampleOrientation}(\vec{e})$ randomly samples $X_o(\vec{e}_i)$ independently with probability density bounded away from zero
%{\em densely} samples~\cite{Lavalle06} $X_o(\vec{e}_i)$ probability one
and $\proc{SampleIK}(p, x_o)$ randomly samples the $(d - 5)$-dimensional space of kinematic solutions independently with probability density also bounded away from zero.
%densely samples the $(d - 5)$-dimensional space of kinematic solutions with probability one.
As a result, \proc{SampleIK} can be used to sample both ${\cal T}(\alpha, \sigma_{\vec{e}_i})$ and ${\cal T}(\sigma_{\vec{e}_i}, \alpha)$ when $x_o = \sigma_{\vec{e}_i}$.
% {\cal T}(\alpha, \sigma_{\vec{e}_i})

\begin{thm} \label{thm:progression}
    \proc{Progression} is probabilistically complete for robustly-feasible extrusion problems.
    \begin{proof}
        We consider a sequence of $m$ events where each event involves both \proc{SampleExtrusion} and \proc{PlanMotion} succeeding given the set of solutions described in definition~\ref{defn:robust}.
        %successfully sampling extrusion parameters using \proc{SampleExtrusion} within the $i$th parameter sets described in definition~\ref{defn:robust} for $i \in \{1, ..., m\}$.
        % (section V-D)
        Because \proc{Progression} is {\em persistent} (section~\ref{sec:persistence}), each search node will be revisited in a finite amount of time.
        As a result, we can ignore the computation in between each revisit.
        For the $i$th event in the sequence, \proc{SampleOrientation} has positive probability of sampling a mode coparameter $\sigma_{\vec{e}_i} \in \Sigma_{\vec{e}_i}$.
        Likewise, \proc{SampleIK} has positive probability of sampling transition configurations $q_{\sigma_{\vec{e}_i}} \in T_{\sigma_{\vec{e}_i}}$ and $q_{\sigma_{\vec{e}_i}}' \in T_{\sigma_{\vec{e}_i}}'$.
        Because \proc{PlanConstrained} and \proc{PlanMotion} are probabilistically complete, for $i$ sufficiently large the probability that they identify a solution is positive.
        %converges to one as $i \to \infty$. 
        As a result, for $i$ sufficiently large, the probability that both \proc{SampleExtrusion} and \proc{PlanMotion} succeed on a given attempt, satisfying the $i$th event, is also positive.
        Thus, the $i$th event will succeed in a finite number of reattempts with probability one, and all $m$ events will succeed in a finite amount of time with probability one.
    \end{proof}
\end{thm}
% If the samplers are correlated this might not actually work
% As long as the \proc{PlanConstrained} and \proc{PlanMotion} procedures are independent

\begin{thm} \label{thm:regression}
    \proc{Regression} is probabilistically complete for robustly-feasible extrusion problems.
    \begin{proof}
        We trivially apply the argument in theorem~\ref{thm:progression} but in the reverse direction from $i \in \{m, ..., 1\}$.
    \end{proof}
\end{thm}

\section{Extrusion Benchmark} \label{appendix:benchmark}
% * results

\begin{figure*}[!ht]
 %\centering
\includegraphics[width=0.32\textwidth]{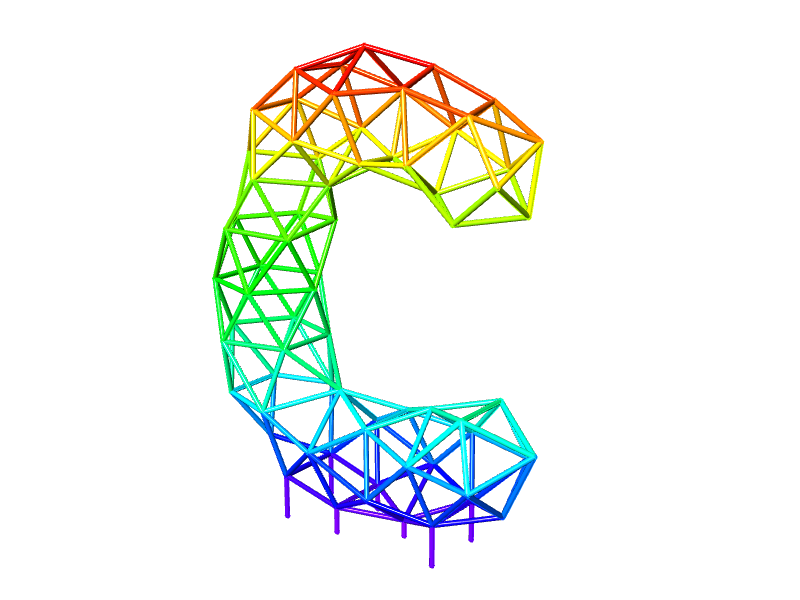}
\includegraphics[width=0.32\textwidth]{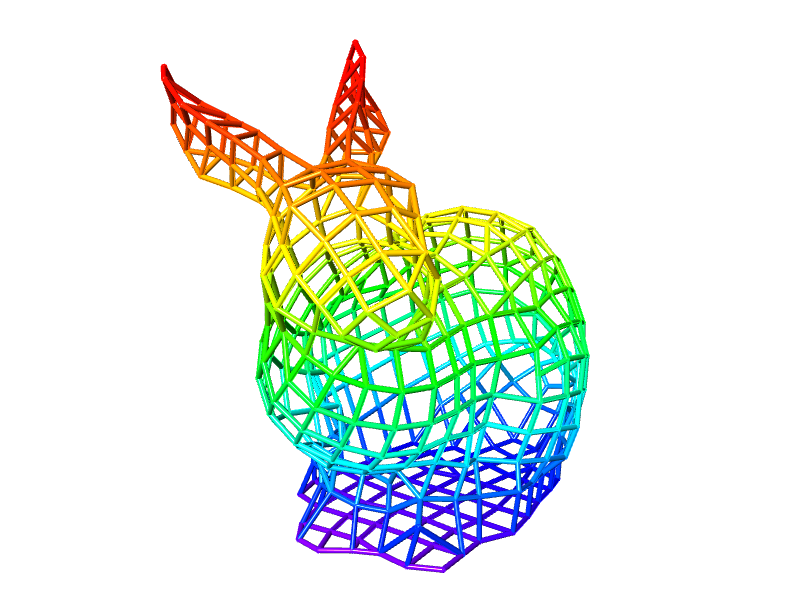}
\includegraphics[width=0.32\textwidth]{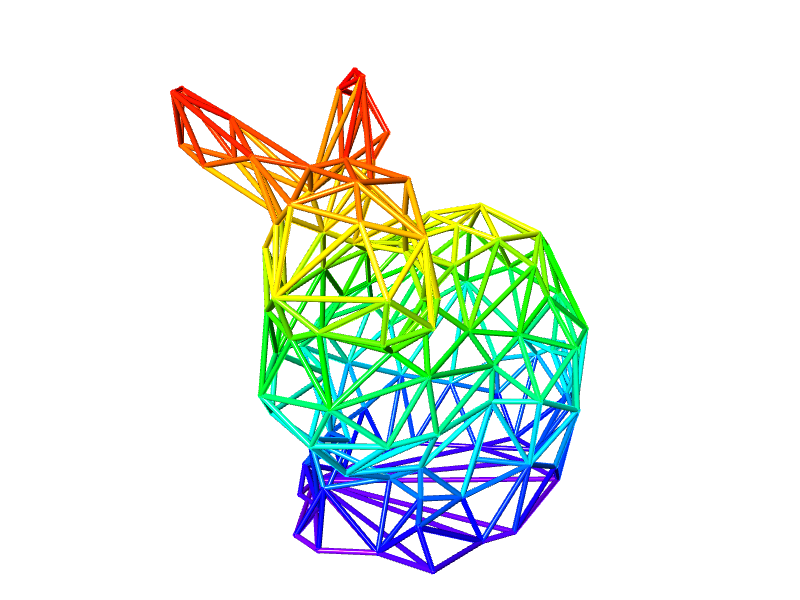}
\includegraphics[width=0.32\textwidth]{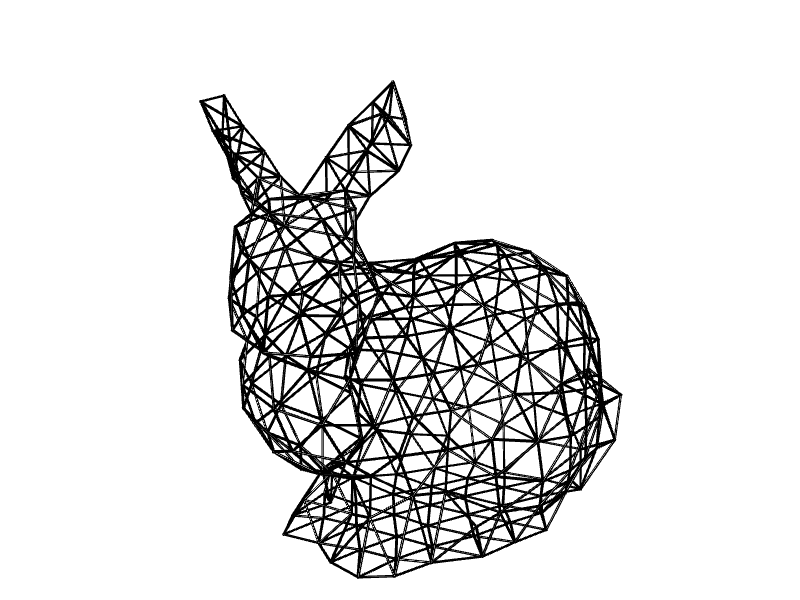}
\includegraphics[width=0.32\textwidth]{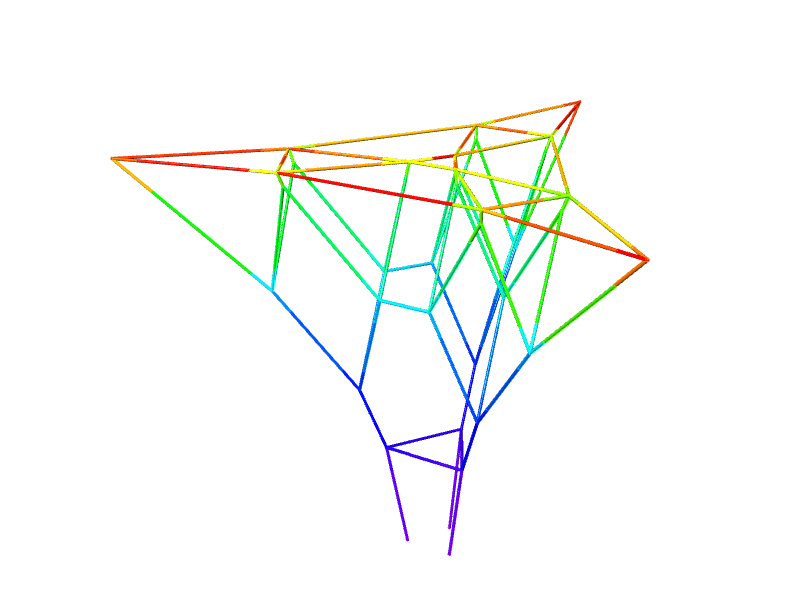}
\includegraphics[width=0.32\textwidth]{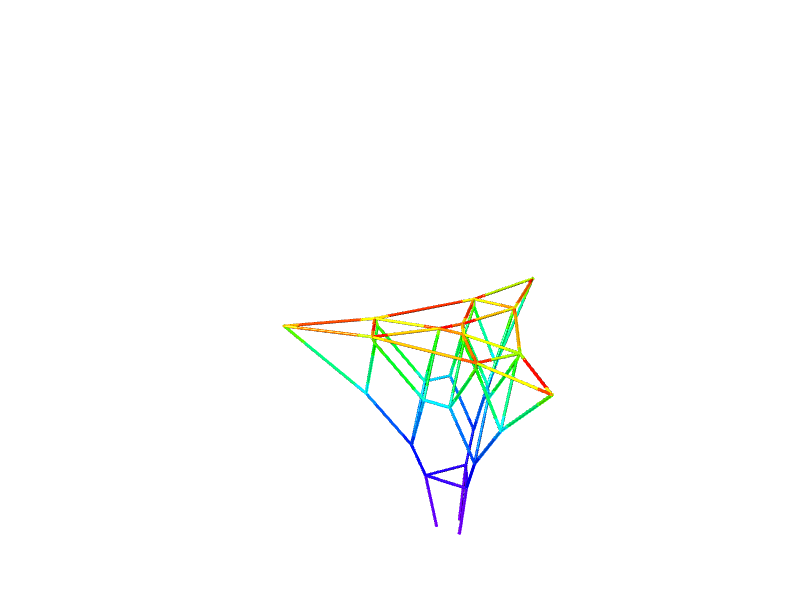}
\includegraphics[width=0.32\textwidth]{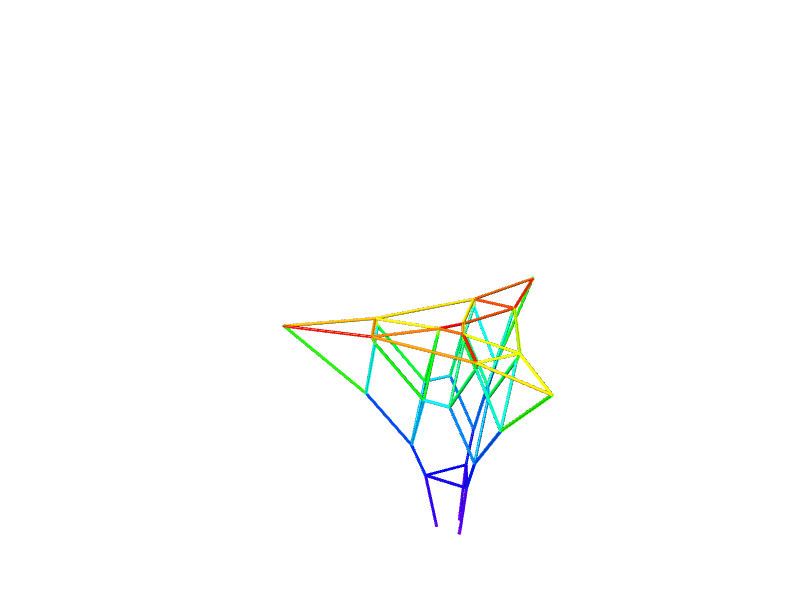}
\includegraphics[width=0.32\textwidth]{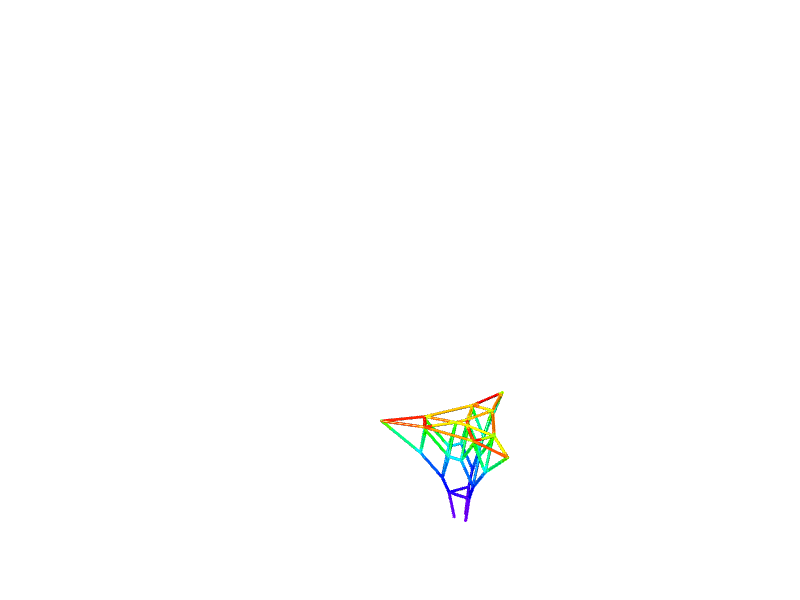}
\includegraphics[width=0.32\textwidth]{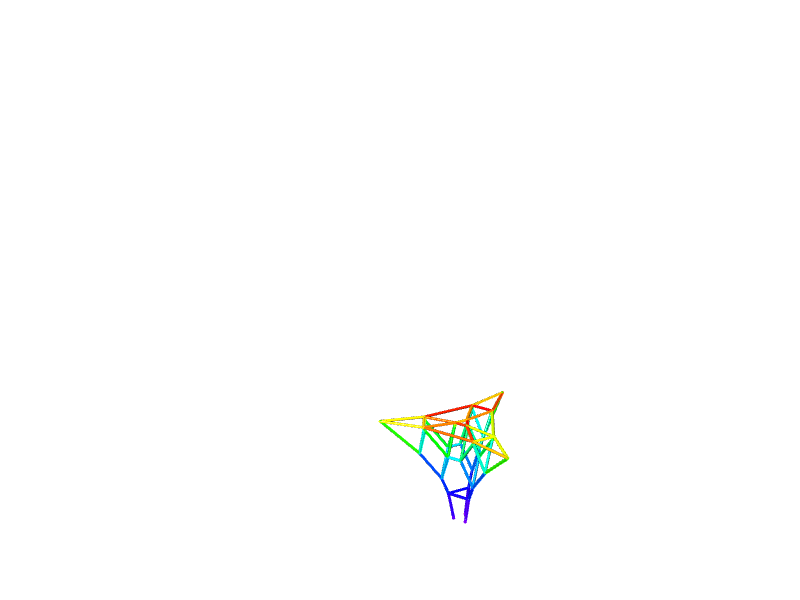}
\includegraphics[width=0.32\textwidth]{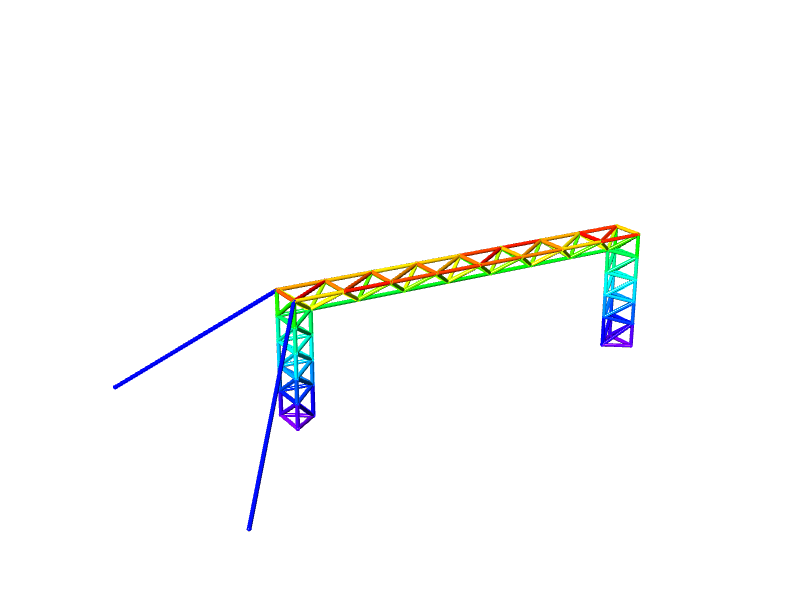}
\includegraphics[width=0.32\textwidth]{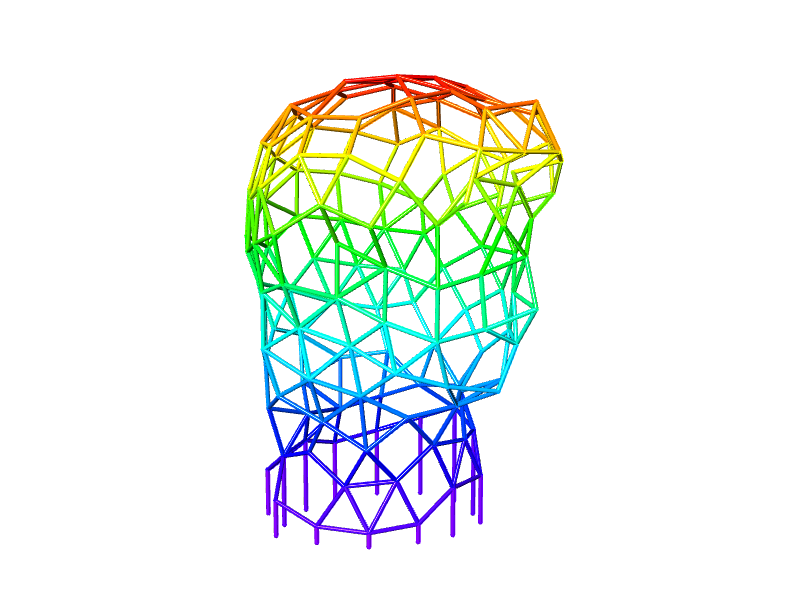}
\includegraphics[width=0.32\textwidth]{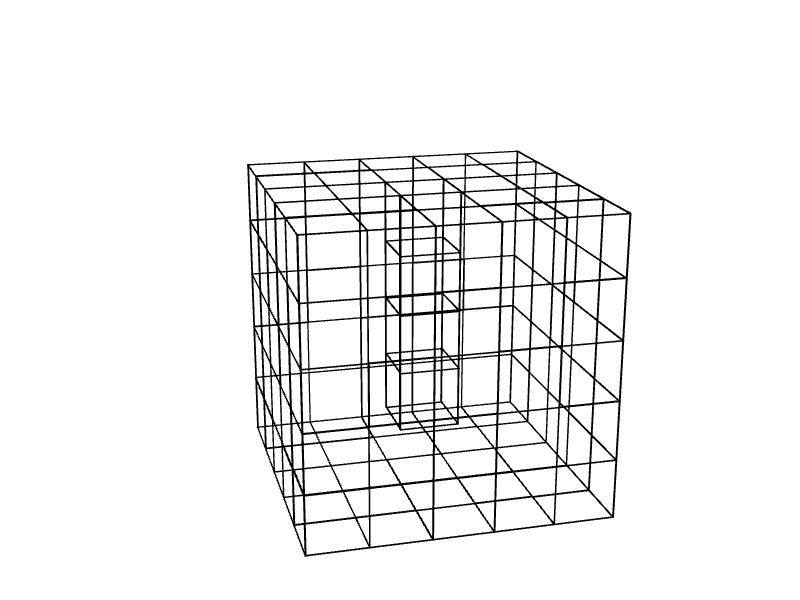}
\includegraphics[width=0.32\textwidth]{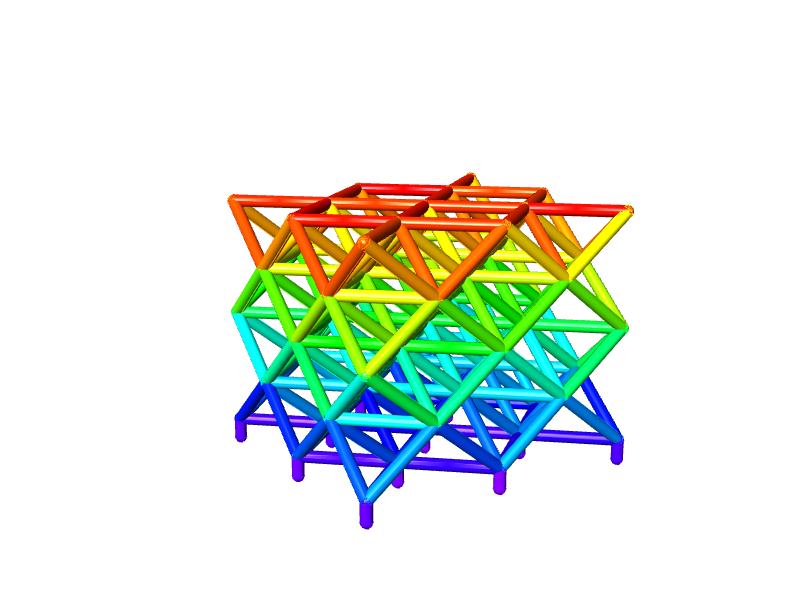}
\includegraphics[width=0.32\textwidth]{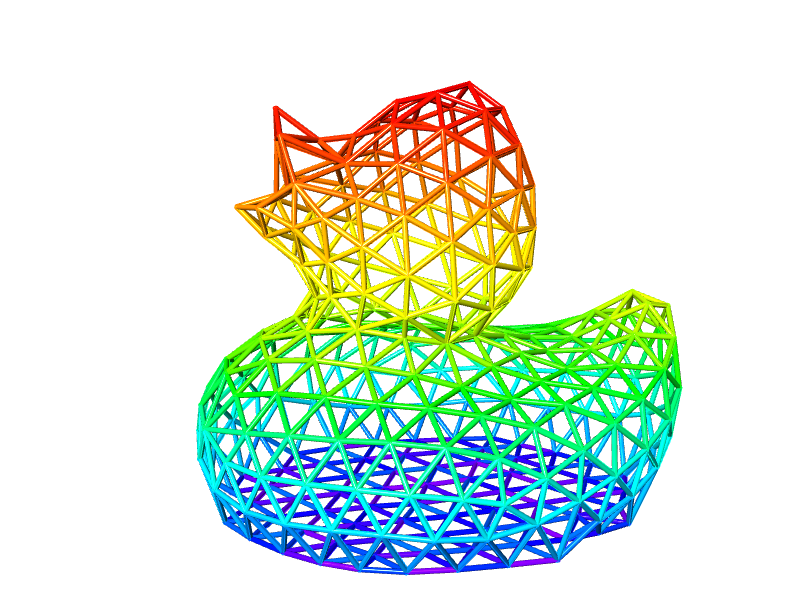}
\includegraphics[width=0.32\textwidth]{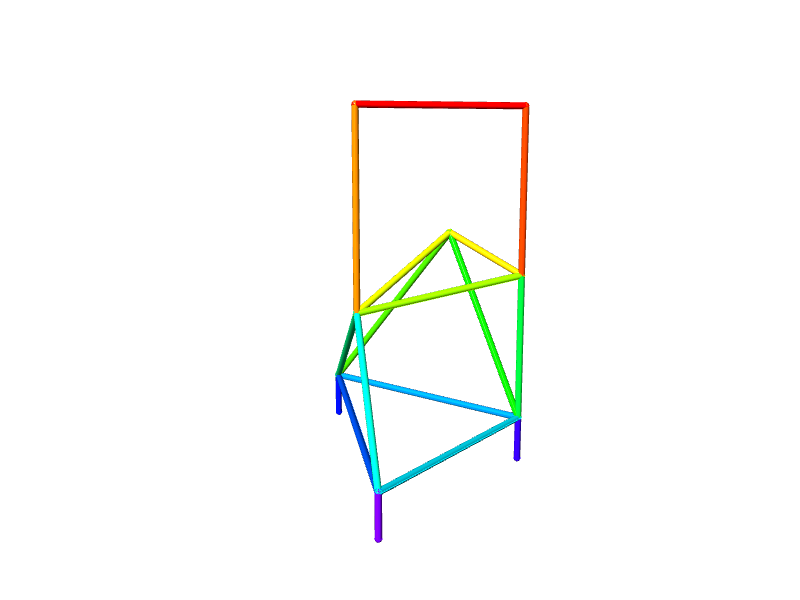}
    \caption{Extrusion Problems}
    \label{fig:instances-1}
\end{figure*}

\begin{figure*}[!ht]
 %\centering
\includegraphics[width=0.32\textwidth]{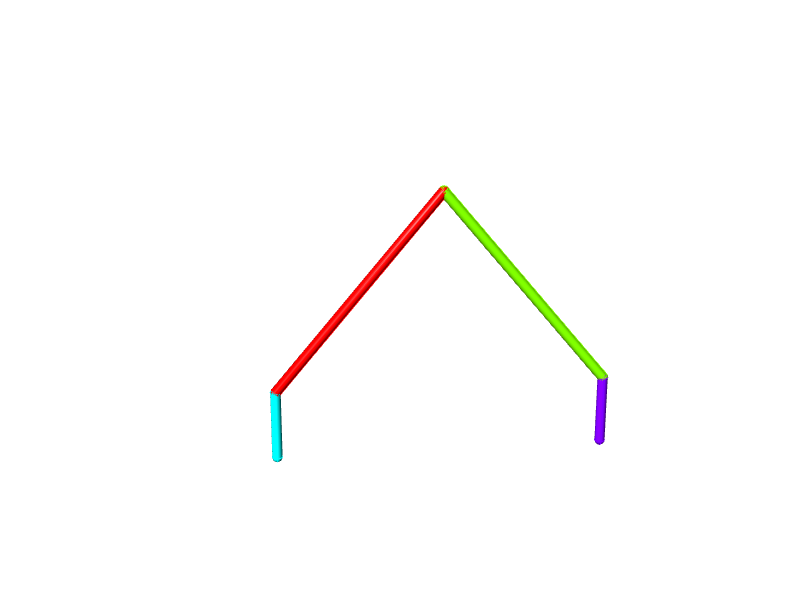}
\includegraphics[width=0.32\textwidth]{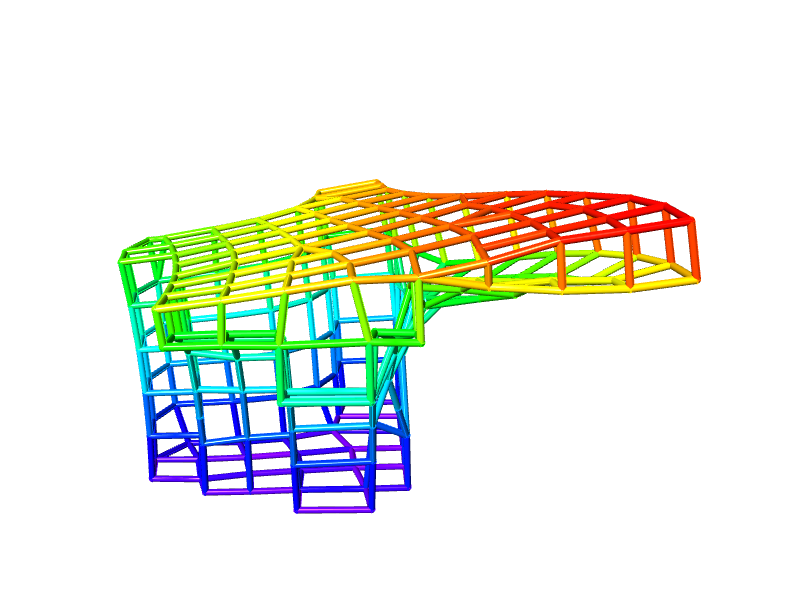}
\includegraphics[width=0.32\textwidth]{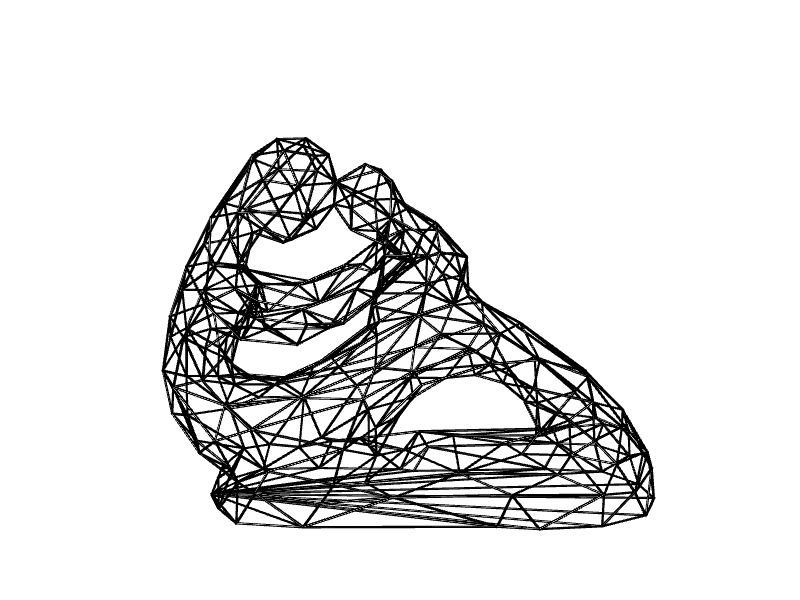}
\includegraphics[width=0.32\textwidth]{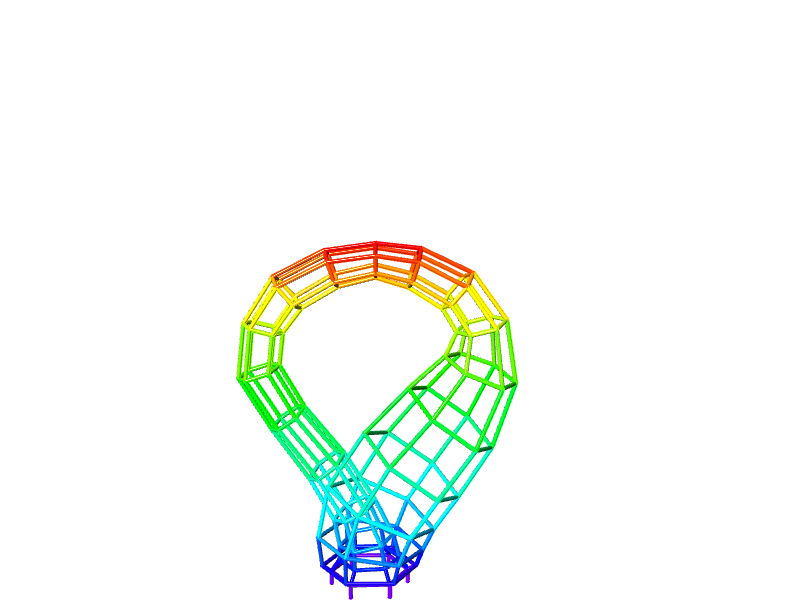}
\includegraphics[width=0.32\textwidth]{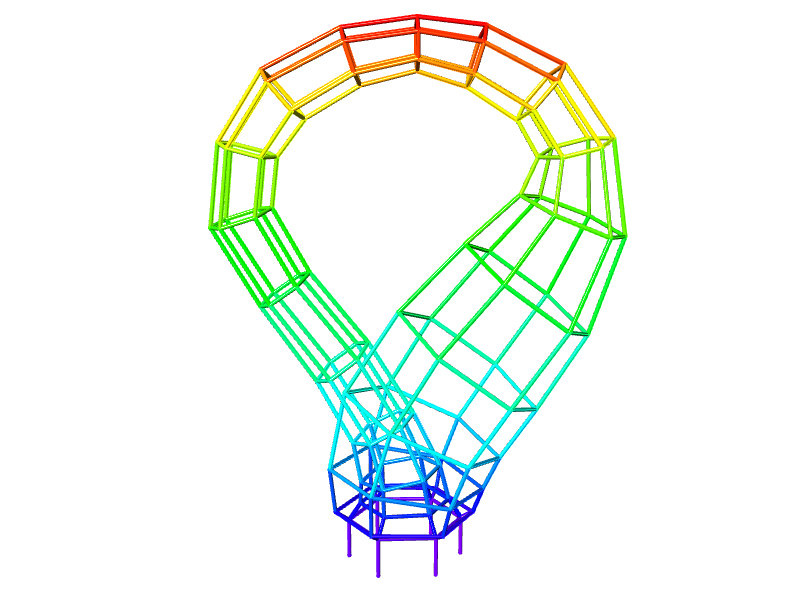}
\includegraphics[width=0.32\textwidth]{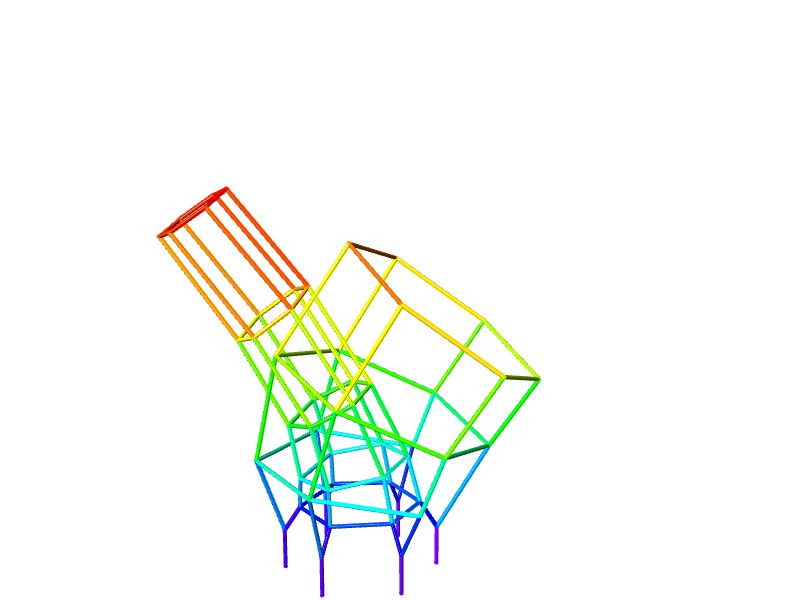}
\includegraphics[width=0.32\textwidth]{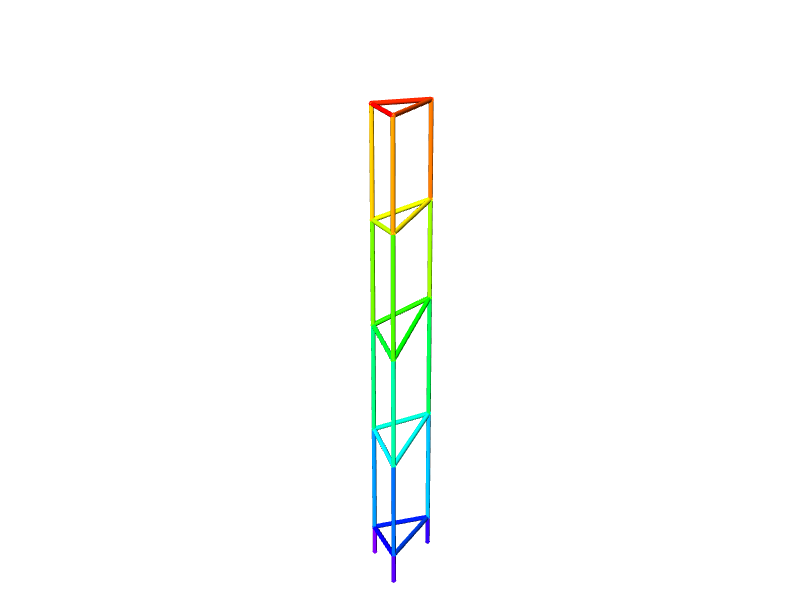}
\includegraphics[width=0.32\textwidth]{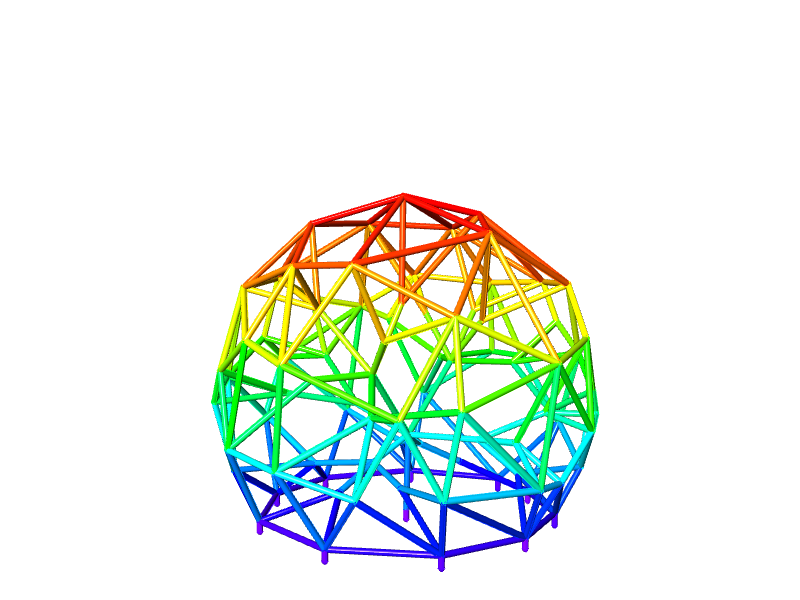}
\includegraphics[width=0.32\textwidth]{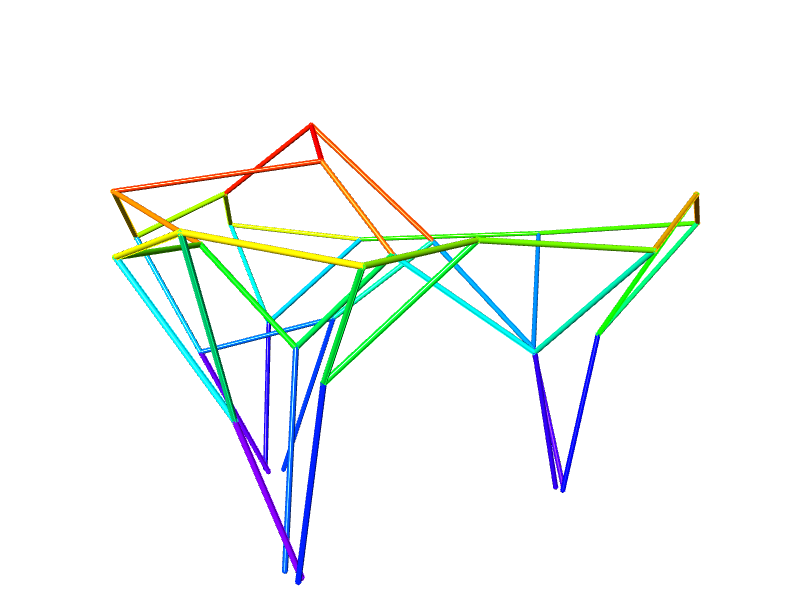}
\includegraphics[width=0.32\textwidth]{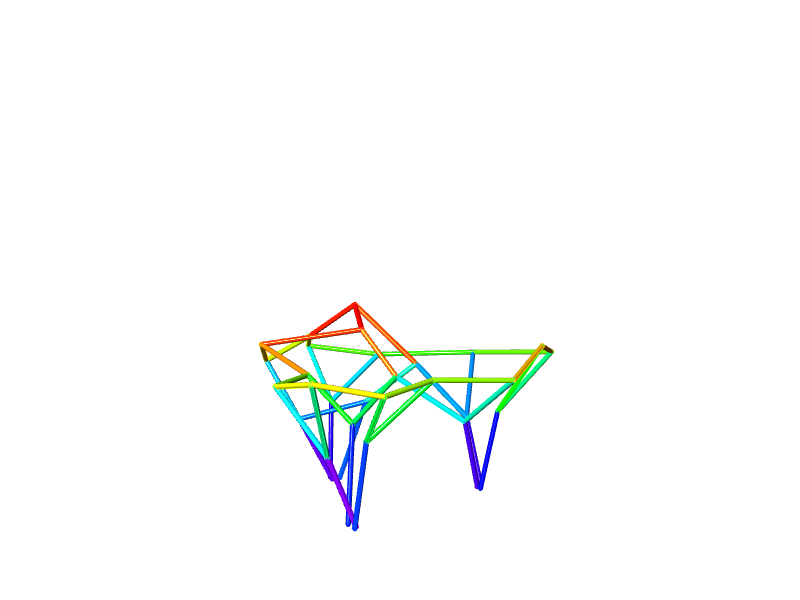}
\includegraphics[width=0.32\textwidth]{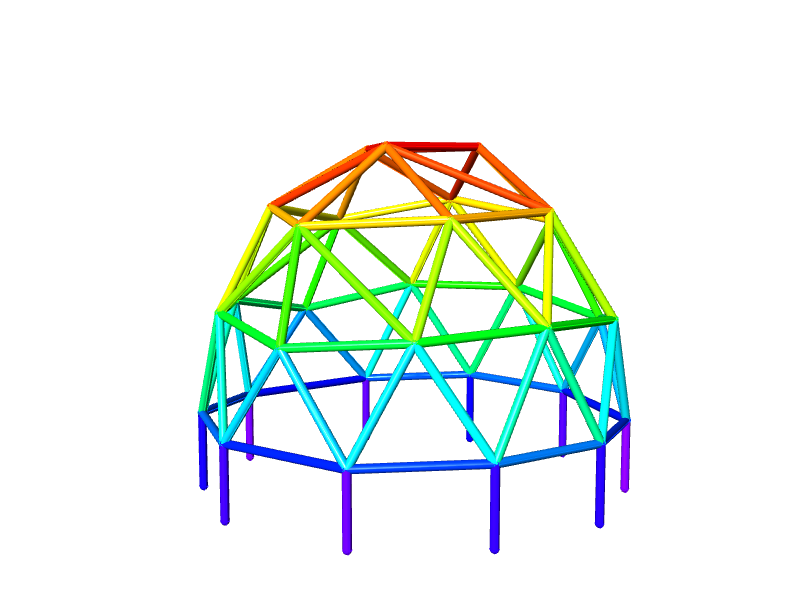}
\includegraphics[width=0.32\textwidth]{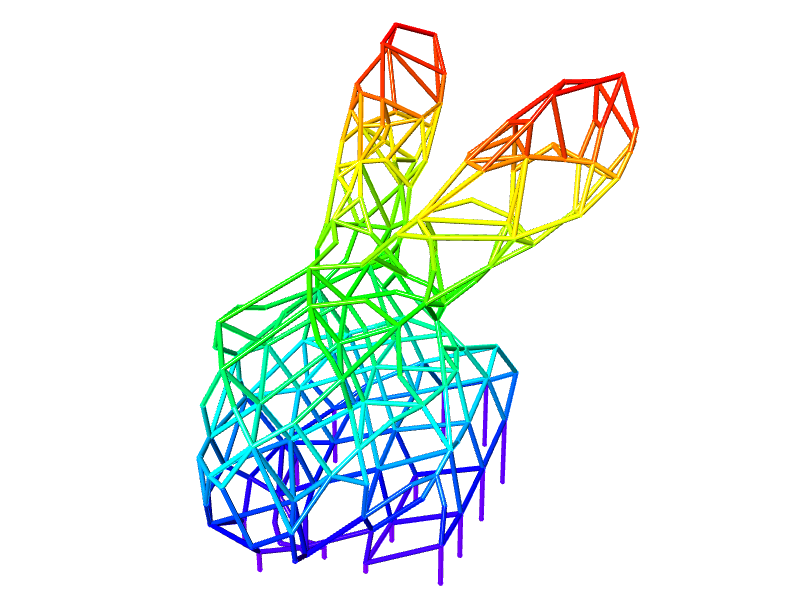}
\includegraphics[width=0.32\textwidth]{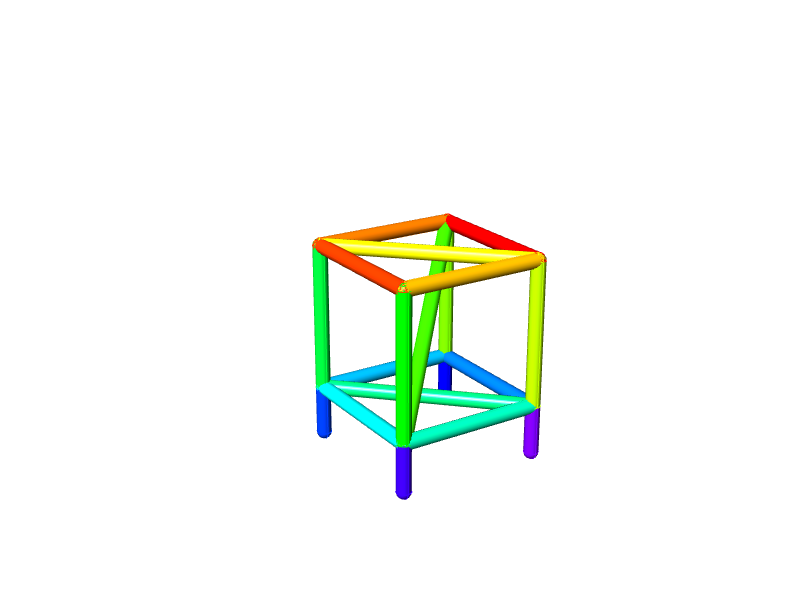}
\includegraphics[width=0.32\textwidth]{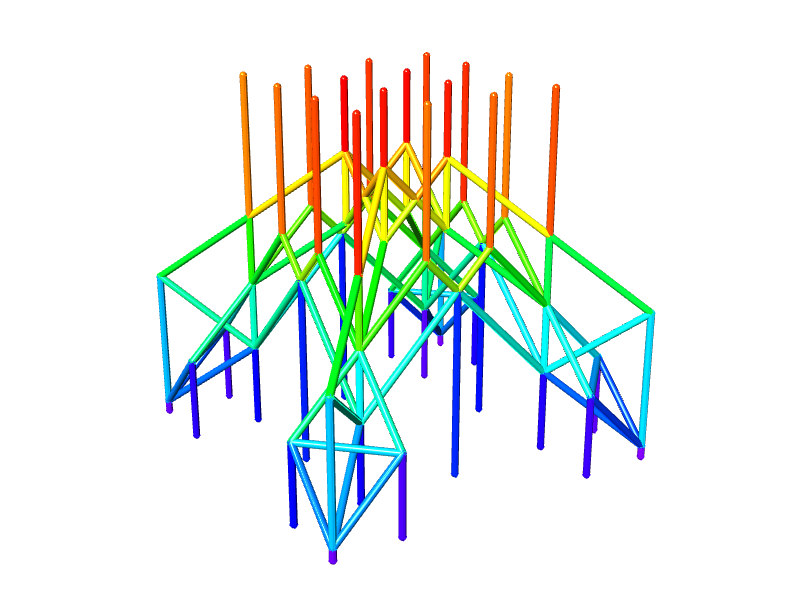}
\includegraphics[width=0.32\textwidth]{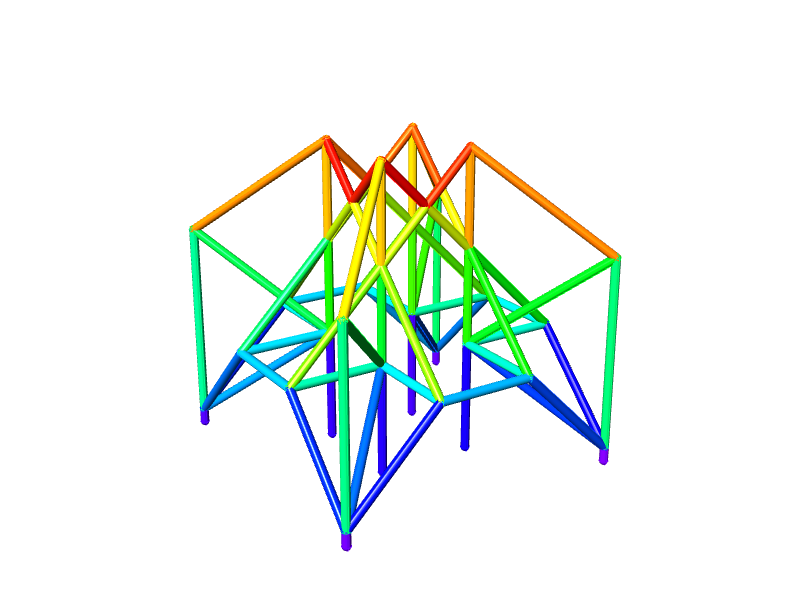}
    \caption{Extrusion Problems}
    \label{fig:instances-2}
\end{figure*}

\begin{figure*}[!ht]
 %\centering
\includegraphics[width=0.32\textwidth]{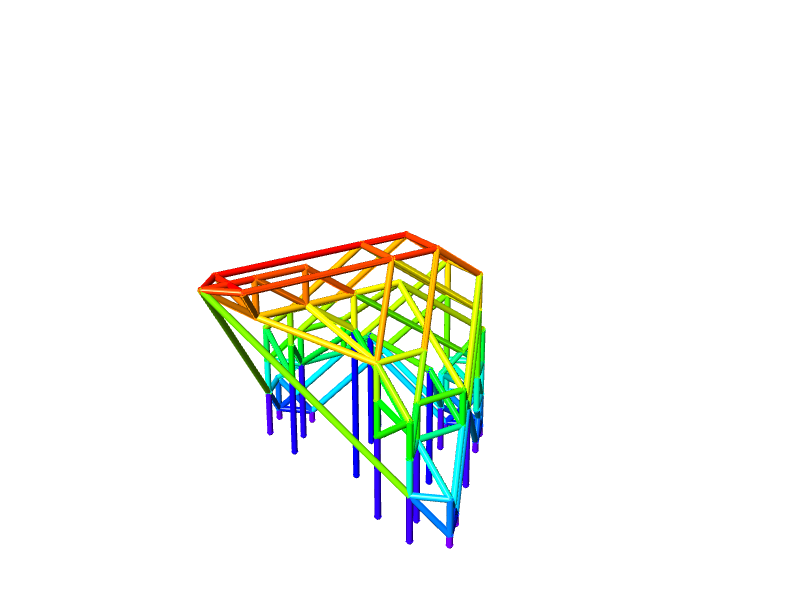}
\includegraphics[width=0.32\textwidth]{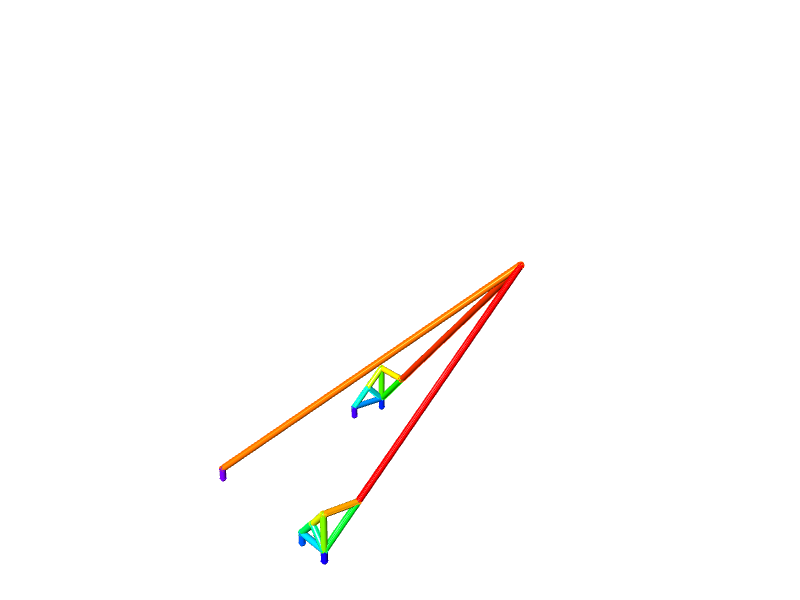}
\includegraphics[width=0.32\textwidth]{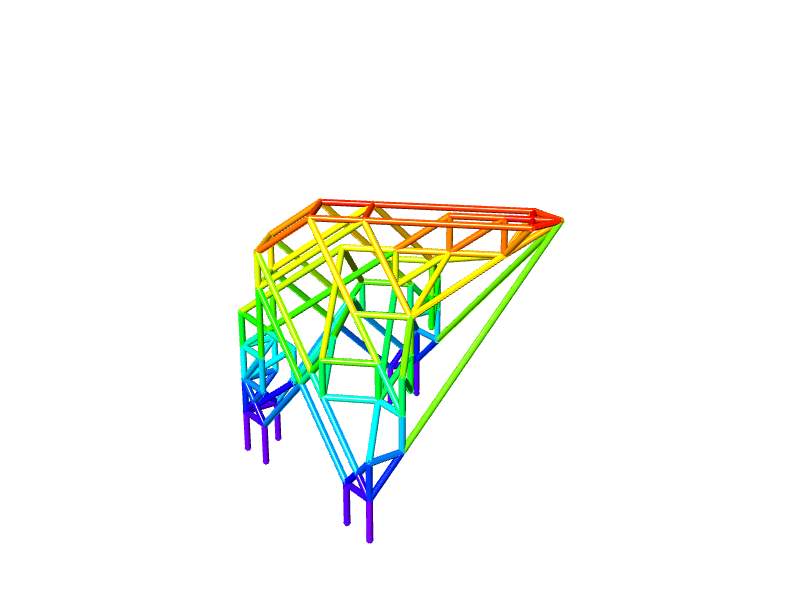}
\includegraphics[width=0.32\textwidth]{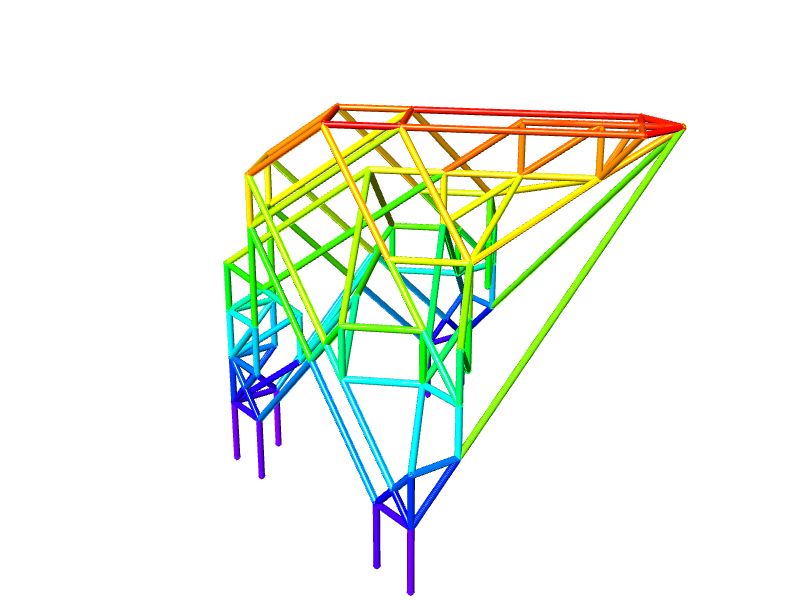}
\includegraphics[width=0.32\textwidth]{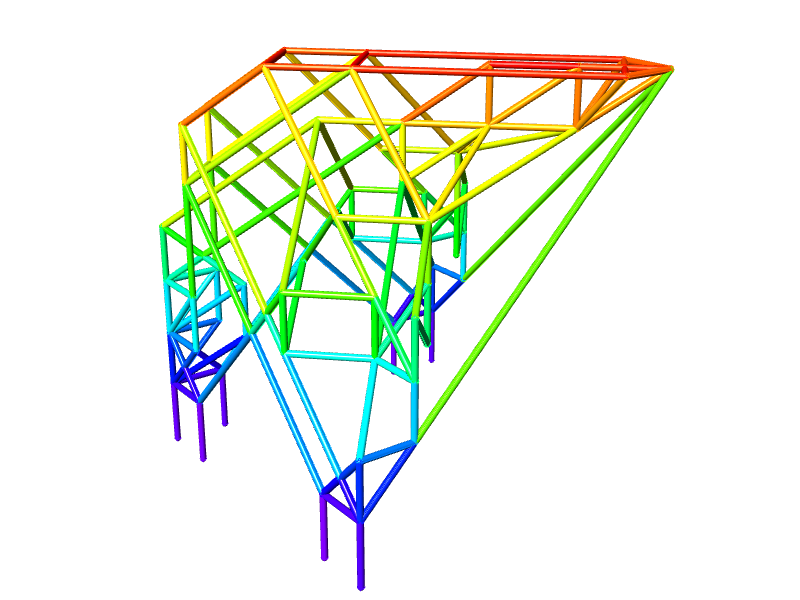}
\includegraphics[width=0.32\textwidth]{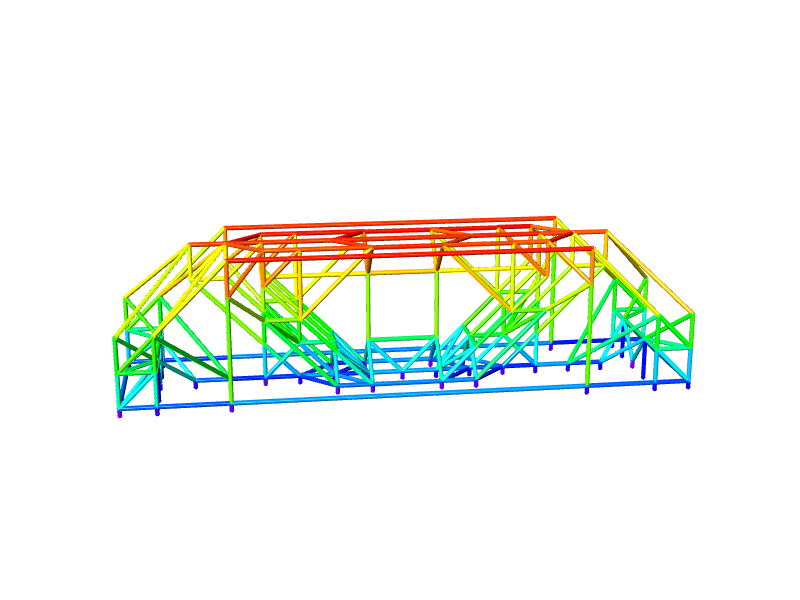}
\includegraphics[width=0.32\textwidth]{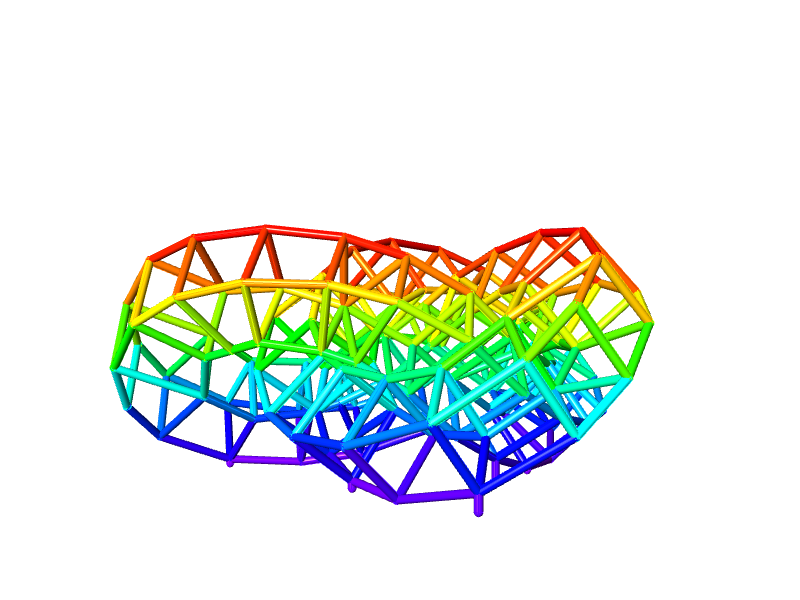}
\includegraphics[width=0.32\textwidth]{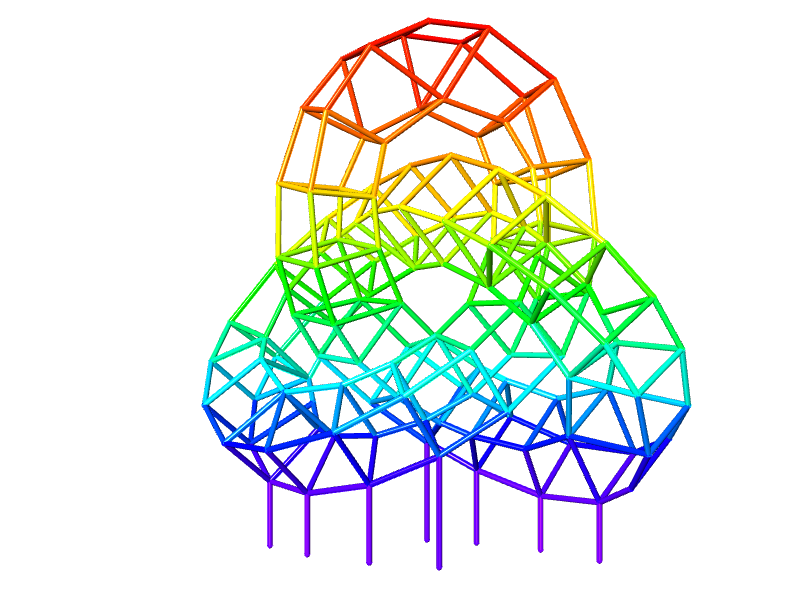}
\includegraphics[width=0.32\textwidth]{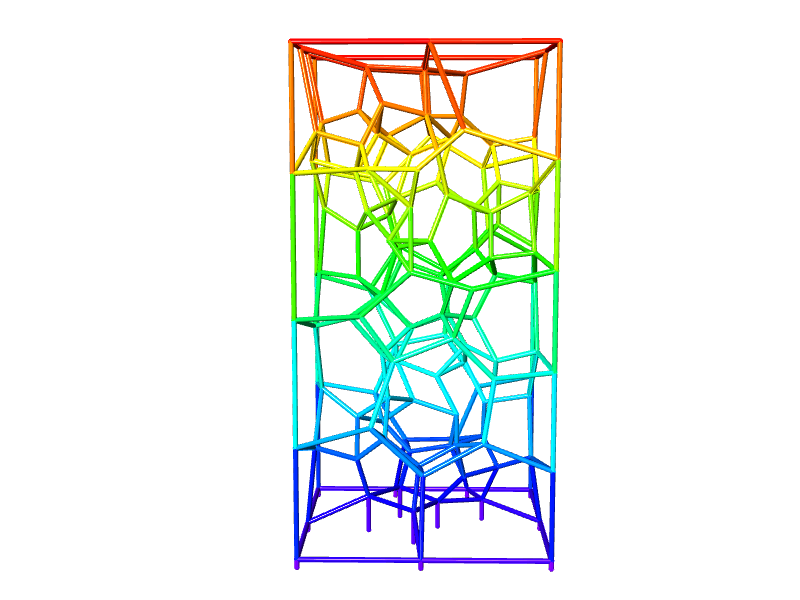}
    \caption{Extrusion Problems}
    \label{fig:instances-3}
\end{figure*}

Figures~\ref{fig:instances-1},~\ref{fig:instances-2}, and~\ref{fig:instances-3} display the extrusion problems that we considered.
For each problem, we ran one trial of \proc{Regression}+{\em StiffPlan} and recorded the extrusion sequence it produced.
For successful trials, elements are colored by their index in a extrusion sequence, where purple elements are printed first and red elements are printed last.
All elements in the structure are black an unsuccessful trial.
Some problems are the result of a linear transformation, such as a rotation or scaling, applied to the same original frame structure.
Other problems are discretized version of the same object but with varying degrees of topological complexity.

% 5*3 + 5*3 + 3*3 + 1 = 40

\newpage

\end{document}